\documentclass{article}

\usepackage[preprint]{neurips_2026}

\usepackage[utf8]{inputenc} 
\usepackage[T1]{fontenc}    
\usepackage{hyperref}       
\usepackage{url}            
\usepackage{booktabs}       
\usepackage{amsfonts}       
\usepackage{nicefrac}       
\usepackage{microtype}      
\usepackage{xcolor}         
\usepackage{amsmath}
\usepackage{wrapfig}
\usepackage{algorithm}
\usepackage{algpseudocode}
\usepackage{float}

\usepackage{caption}
\usepackage{graphicx} 

\title{Graph Hierarchical Recurrence \\ for Long-Range Generalization}

\author{%
  Stefano Carotti$^{1,2}$ \quad
  Marco Pacini$^{2}$ \quad
  Alessio Gravina$^{3}$ \And
  Davide Bacciu$^{3}$ \quad
  Bruno Lepri$^{2}$ \quad
  Sebastiano Bontorin$^{2}$%
}

\newcommand{\rev}[1]{{\color{black} #1}}

\begin{document}

\maketitle

{\let\thefootnote\relax\footnotetext{$^{1}$Department of Computer Science, University of Trento, Italy.\  $^{2}$Fondazione Bruno Kessler, Italy.\ $^{3}$Department of Computer Science, University of Pisa, Italy.\ Correspondence: \texttt{scarotti@fbk.eu}.}}


\begin{abstract}

Graph Neural Networks and Graph Transformers have become central to graph learning, combining expressive representation learning with sample-efficient inductive biases.
Yet they remain fundamentally limited when predictions depend on correlations between distant graph regions.
We address this limitation with Graph Hierarchical Recurrence (GHR), a novel framework that jointly operates on the input graph and a pooled hierarchical abstraction.
We also show that existing models degrade more sharply under out-of-range generalization, where test instances require interactions across distances exceeding those observed during training. Despite its minimal design, GHR consistently strengthens every tested message-passing backbone, yielding robust performance on long-range dependencies and particularly pronounced gains in out-of-range regimes.
Across a broad suite of long-range benchmarks, GHR achieves state-of-the-art or competitive results on multiple tasks, establishing hierarchical recurrence as an effective mechanism for extending graph models beyond their observed interaction range.

\end{abstract}

\section{Introduction}
\label{intro}

\begin{figure}[h]
\begin{center}
\includegraphics[width=1.0\linewidth]{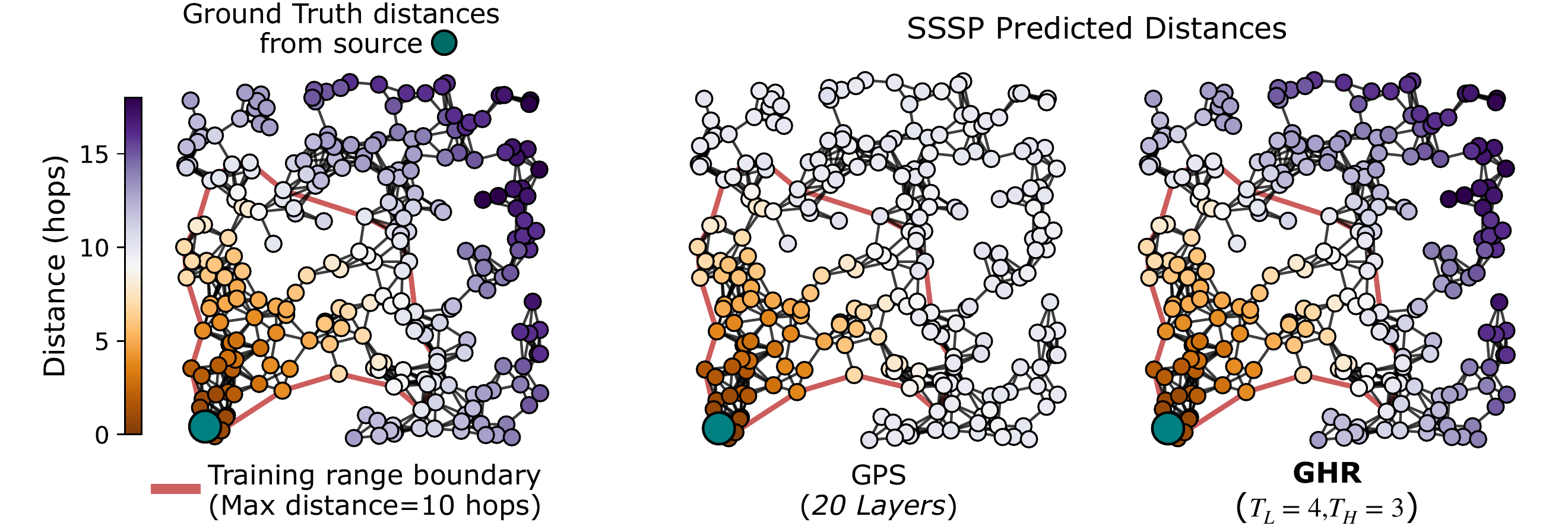}
\end{center}
\caption{\textbf{Out-of-Range Generalization for Graph Hierarchical Recurrence.} Predicted distances on Single Source Shortest Path (SSSP) task from a sample source node in a Random Geometric Graph (RGG). We compare \rev{GPS} (20 layers) with our model GHR ($T_{L}=4$, $T_{H}=3$), both trained on distances of up to 10 hops. While both \rev{GPS} and GHR accurately predict on in-range distances, only GHR achieves out-of-range generalization.}
  \label{H}

\end{figure}
Graph Neural Networks (GNNs) \citep{NN4G, scarselli2009gnn} are a standard framework for learning on graph-structured data.
Most architectures rely on Message-Passing Neural Networks (MPNNs) \citep{gilmer2017mp}, where each layer expands the receptive field by one hop. A model shallower than the graph diameter therefore cannot exchange signals between every pair of nodes, a phenomenon known as under-reaching \citep{barcelo2020logical, errica2025adaptive}. Fixed-depth MPNNs are consequently ill-suited to global tasks whose required propagation depth grows with the graph diameter~\citep{loukas2020what}. Adding layers extends the receptive field but exacerbates over-squashing \citep{alon2021on, topping2022understanding, diGiovanniOversquashing, mishayev2025shortrange}, over-smoothing \citep{oono2020graph, cai2020note}, and vanishing gradients \citep{arroyo2025}. Graph Transformers mitigate these propagation limits through global attention \citep{rampasek2022GPS}, but their quadratic complexity becomes prohibitive on large graphs.
Several linear-complexity alternatives shorten communication paths. Graph rewiring \citep{topping2022understanding, gutteridge2023drew} and virtual nodes \citep{southern2025understanding} introduce additional connections, at the cost of altering the original topology \citep{arnaiz2022diffwire}.
Hierarchical pooling \citep{ying2018hierarchical, bianchi2023the, lachi2025expressive} enlarges the receptive field by coarsening the graph through clustering and feature aggregation. Although this alleviates under-reaching, unpooling and reconstruction can degrade the node-level information required by many long-range tasks. Moreover, fixed depth models still permit only a fixed number of propagation steps.
Weight-shared recurrent architectures \citep{scarselli2009gnn, gravina_swan, tang2020} decouple computation depth from parameter count and enforce a step-invariant update. In neural algorithmic reasoning, this inductive bias supports extrapolation to longer reasoning horizons \citep{Velickovic2020Neural, zhou_ood_2022}, typically when the architecture is aligned with the target algorithm so that aggregation encodes its nonlinear structure and learned components need to represent only linear functions \citep{xu2021how}. Such alignment, however, requires an explicit algorithmic specification of the task. Even then, spanning large graph diameters still requires proportionally many message-passing steps and inherits the long-range limitations above.

Literature on extrapolation in graph learning has focused on generalization to larger graphs \citep{xu2021how, yehudai2021local} and longer reasoning horizons. In this paper, we show that generalization also depends on interaction distances observed during training. We formalized this distinction through \textbf{in-range generalization}, where test instances require propagation over distances observed during training, and \textbf{out-of-range generalization}, where test instances require propagation beyond the training interaction range  (Figure~\ref{H}).
Out-of-range cases may result from sampling bias or distribution shifts, but can also arise entirely in-distribution through finite sampling, for example, when a shortest-path instance graph drawn from the same graph-size distribution contains paths longer than any observed during training. We find that many established models perform well within the training interaction range yet degrade sharply beyond it (Table~\ref{tab:echo-synth-oor}).

To address these limitations, we introduce Graph Hierarchical Recurrence (GHR), a framework that couples recurrent computation on the input graph, with parallel message passing on a coarsened graph, inspired by the Hierarchical Reasoning Model \citep{wang2025hierarchical}. Recurrence provides step-invariant updates and allows trained layers to be reused for additional test-time iterations, while the coarsened graph supplies a shorter route for long-range communication. 
GHR preserves the fine-graph state across iterations through a residual connection and uses the coarse-graph state only as an additional input to each fine-graph update.
This design creates a parallel route through the hierarchy without downsampling or mixing away the fine-grained representation. Unlike architectures aligned to a specific target algorithm, GHR imposes no algorithmic constraint and can augment any message-passing backbone, including on tasks without an explicit algorithmic description.

Our contributions are summarized as follows:
\begin{itemize}
    \item We introduce GHR, a graph learning framework that integrates recurrent computation with coupled message passing over the original graph and a coarsened representation through iterative pooling and unpooling (Section~\ref{model}).
    \item We formalize in-range and out-of-range generalization and show, in controlled experiments, that widely used baselines fail in out-of-range, whereas combining hierarchical coarsening with recurrence improves performance in this regime (Section~\ref{sec:Ablations}).
    \item We evaluate GHR on established long-range benchmarks, achieving state-of-the-art or competitive results and consistently improving every corresponding flat model across all message-passing backbones considered (Section~\ref{sec:benchmarks}).

\end{itemize}
\vspace{-8pt}

\section{Preliminaries}
\label{background}

\paragraph{Graphs and Features.} 

Let $G = (V, E, \mathbf{X}, \mathbf{e})$ be an attributed graph, where $V$ is a finite set of nodes and $E \subseteq \{\{i,j\} : i,j \in V,\, i \neq j\}$ is the edge set.
Each node $u \in V$ is associated with a feature vector $\mathbf{x}_u \in \mathbb{R}^{d_n}$, and the matrix collecting all node features is denoted by $\mathbf{X} \in \mathbb{R}^{|V| \times d_n}$.
Each edge $\{i,j\} \in E$ is associated with an feature $\mathbf{e}_{ij} \in \mathbb{R}^{d_e}$, and we denote the collection of edge features by $\mathbf{e} = \{\mathbf{e}_{ij} : \{i,j\} \in E\}$ and the neighborhood of a node $i$ as $\mathcal{N}_i := \{j \in V : \{i,j\} \in E\}$.

\vspace{-4pt}
\paragraph{Message Passing.}
We define a message-passing layer as a function that processes the features on a graph $G$ and updates node features by aggregating information from neighboring nodes and incident edges.
More precisely, for a graph with node features $\mathbf{X} \in \mathbb{R}^{|V| \times d_n}$ and edge features $\mathbf{e}_{ij} \in \mathbb{R}^{d_e}$, each layer produces an updated node feature $\mathbf{x}'_i \in \mathbb{R}^{d'_n}$ for each node $i \in V$ via:
\begin{equation}
\label{eq:MP}
    \mathbf{x}'_i =
    \mathrm{MP}_{W}
    \left(
        \mathbf{x}_i,
        \{\!\{ \mathbf{x}_j \}\!\}_{j \in \mathcal{N}_i},
        \{\!\{ \mathbf{e}_{ij} \}\!\}_{j \in \mathcal{N}_i}
    \right),
\end{equation}
where $\mathrm{MP}_{W}$ is a learnable function with parameters $W$.
Here, $\{\!\{ \cdot \}\!\}$ denotes a multiset, i.e., a collection where elements may appear with multiplicity.
For simplicity, we will often write $\mathrm{MP}_W(\mathbf{X}, \mathbf{e})$.
\vspace{-12pt}
\paragraph{Hierarchy: Graph pooling and coarsening.}
To expand the receptive field of the model and process graph data at a coarser scale, we introduce a two-level hierarchy induced by a graph pooling map. In general, graph \emph{pooling} maps a graph $G = (V, E, \mathbf{X}, \mathbf{e})$ to a typically smaller graph $G' = (V', E', \mathbf{X}', \mathbf{e}')$, while \emph{unpooling} maps features from $G'$ back to the original graph structure using cluster assignments. We denote these maps by $\mathrm{Pool} : G \mapsto G'$ and $\mathrm{Unpool} : G' \mapsto G$, with $\mathrm{Unpool} \circ \mathrm{Pool}(G) = (V, E, \mathbf{X}'', \mathbf{e}'')$, so that the underlying graph structure $(V,E)$ is preserved, while node and edge features may be modified.
The pooled graph $G'=(V',E')$ is constructed by applying a clustering algorithm to $G=(V,E)$ once, assigning each node to a super-node. By a slight abuse of notation, we also denote this assignment by $\mathrm{Pool}$. 
In the GHR framework, $\mathrm{Pool}$ acts as a graph quotient map, i.e., $V'$ is a partition of $V$ and $E'$ contains an edge between two super-nodes whenever any pair of their members is connected in $E$ \citep{bader2013graph}. As a general approach we adopt topological clustering such as Graclus \citep{dhillon2007weighted}, while for spatially embedded graphs, the underlying metric can also be exploited. This is the case for geometric block pooling that we employ in the LRIM benchmark, in Section \ref{experiments}.  Both algorithms and the relation to a flat recurrent model, are further described in Appendix~\ref{pooling-examples}. \\
In $\mathrm{Pool}$, node features within each cluster are aggregated into $\mathbf{X}'$ using a task-dependent pooling operator (e.g., sum, mean, or max) \citep{grattarola2022understanding}. 
In $\mathrm{Unpool}$, the features of nodes in $V'$ are broadcast to the nodes in $V$ following the cluster assignments.
In our setting, we instantiate this construction with two levels: the \emph{low-level graph} $G_L$ and the \emph{high-level graph} $G_H := \mathrm{Pool}(G_L)$. In the following, the subscript $H$ denotes quantities associated with $G_H$, such as $\mathbf{X}_H$ and $\mathbf{e}_H$, while the subscript $L$ denotes the corresponding quantities associated with $G_L$.


\section{The Graph Hierarchical Recurrence Framework}
\label{model}
\begin{figure}
\begin{center}
  \includegraphics[width=0.85\linewidth]{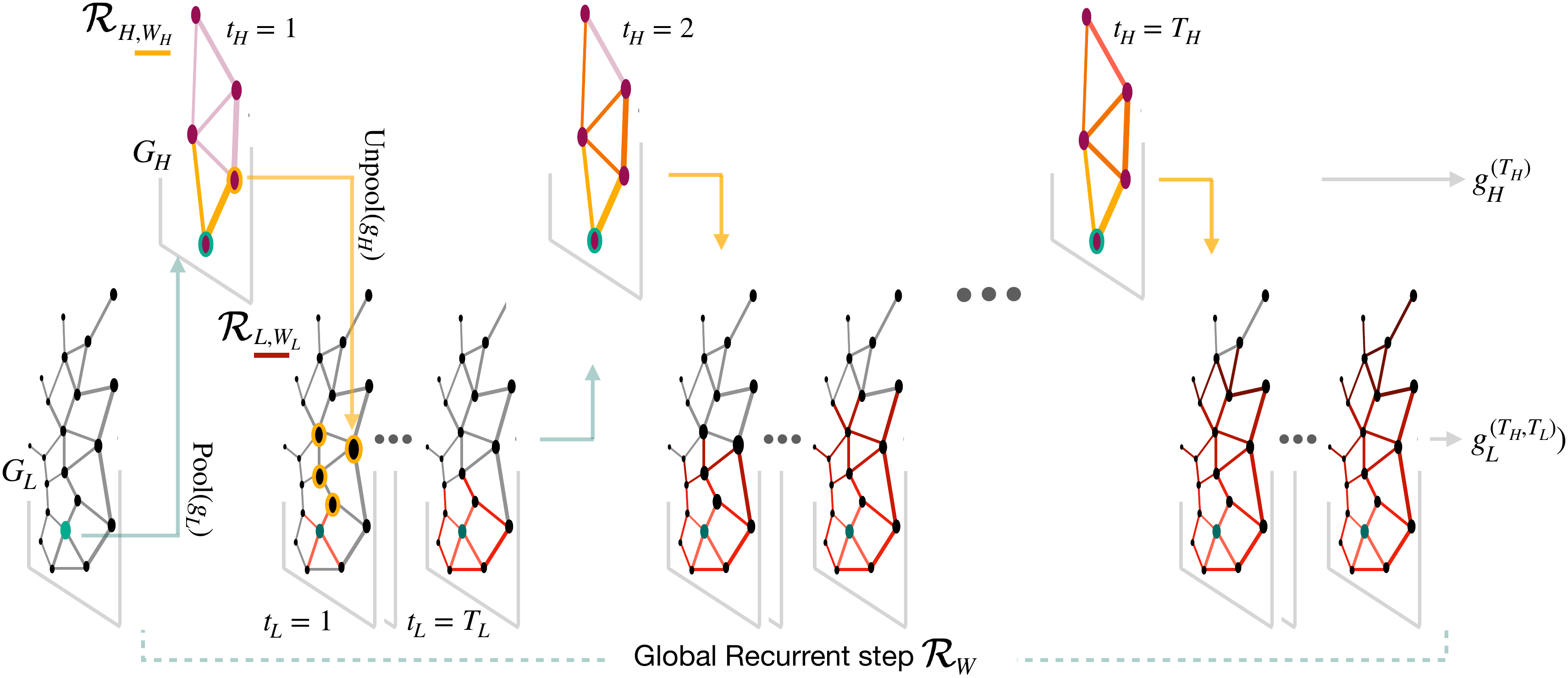}
\end{center}
\caption{\textbf{Visualization of the global recurrent step $\mathcal{R}_{W}$.} Visual representation of the nested recurrent scheme as described in Algorithm \ref{alg:global_step}. $\mathcal{R}_{L,W_{L}}$ and $\mathcal{R}_{H,W_{H}}$ propagate nodes' information across the low-level graph $G_L$ and the high level abstraction $G_H$. Message passing is interleaved between two layers in a nested weight-sharing scheme.}
  \label{fig:unrolling}
\vspace{-8pt}
\end{figure}

In this section, we introduce our framework, Graph Hierarchical Recurrence (GHR),
with the aim of alleviating the out-of-range generalization problem illustrated in Figure~\ref{H} and Figure~\ref{fig:GHR_vs_GPS}. In particular, we define \textbf{in-range (IR)} generalization as the ability to generalize to instances requiring interaction distances seen during training, and \textbf{out-of-range (OOR)} generalization as the ability to extend this capability to interaction distances longer than those observed during training. The framework of GHR consists of a preprocessing step, in which the high-level graph $G_H = \mathrm{Pool}(G_L)$ is generated and the input features are encoded, followed by the iterative application of a learnable global recurrent step $\mathcal{R}_{W}$ (Section~\ref{section:reasoning_step}).
Specifically, each global step comprises two parallel propagation streams (as shown in Figure~\ref{fig:unrolling}). The low-level message passing operates on the original graph and captures fine-grained local interactions between nodes. The high-level message passing operates on a pooled version of the graph and enables efficient long-range communication. These two streams exchange information via an iterative pooling-unpooling scheme, allowing representations at different scales to influence one another. A diagram of the architecture is shown in Figure~\ref{fig:full_architecture} and Figure~\ref{fig:global_recurrent_step}. 

\paragraph{Generation of $G_H$ and input encoding.}
Let $G_L = (V_L, E_L, \mathbf{X}_L, \mathbf{e}_L)$ be the input graph and $G_H = (V_H, E_H, \mathbf{X}_H, \mathbf{e}_H) := \mathrm{Pool}(G_L)$ its high-level counterpart, computed once per graph in a pre-processing step (Section~\ref{background}). In contrast to other hierarchical pooling methods \citep{ying2018hierarchical}, GHR retains the topology of the original graph throughout the computation. 
Node and edge features of both levels are projected into an $m$-dimensional latent space by learnable linear maps. We denote the encoded node features by $f_L = W_n \mathbf{X}_L \in \mathbb{R}^{|V_L| \times m}$, and keep $\mathbf{e}_L$ and $\mathbf{e}_H$ for the encoded edge features. This decouples the input dimensions from the parametrization of the recurrent step, so that the hyperparameters of $\mathcal{R}_W$ can be chosen independently (details in Appendix~\ref{app:input-encoding}).

\paragraph{Global recurrent step and hidden states.}
The \textit{global recurrent step} of GHR, $\mathcal{R}_W$, is defined as
\begin{equation}
    \label{eq:rs}
    \left(h_L^{(r)}, h_H^{(r)}\right)
    = \mathcal{R}_W\!\left(f_L, \mathbf{e}_L, \mathbf{e}_H, h_L^{(r-1)}, h_H^{(r-1)}\right),
    \quad r = 1, \dots, R.
\end{equation}
The encoded features $f_L$, $\mathbf{e}_L$ and $\mathbf{e}_H$ are injected unchanged at every iteration, whereas the hidden states $h_L^{(r)} \in \mathbb{R}^{|V_L| \times m}$ and $h_H^{(r)} \in \mathbb{R}^{|V_H| \times m}$ are updated recurrently and constitute the evolving state of the model. The hidden states $h_L^{(0)}$ and $h_H^{(0)}$ are obtained from a learnable vector broadcast to all nodes  (Appendix~\ref{app:input-encoding}).
The two-level recurrent structure of the global recurrent step $\mathcal{R}_W$ is described in detail in Section~\ref{section:reasoning_step}.
A diagram of the iterative application of $\mathcal{R}_W$, together with the computation of the intermediate loss across the $R$ steps, is provided in Figure~\ref{fig:full_architecture} in Appendix~\ref{sec:additional_arch}.

\subsection{The Global Recurrent Step}
\label{section:reasoning_step}

As specified at the beginning of Section~\ref{model}, the core building block of GHR is the global recurrent step $\mathcal R_W$. Figure~\ref{fig:unrolling} illustrates that $\mathcal R_W$ performs joint propagation at two levels of resolution: low-level message passing on the original graph $G_L$, and high-level message passing on the coarser pooled graph $G_H$. This high-level stream alleviates long-range propagation issues exploiting the contraction of distances on $G_{H}$, lowering the number of message-passing steps needed to span the graph~\citep{diGiovanniOversquashing}, as we discuss in Section ~\ref{sec:contraction}.

\begin{wrapfigure}{r}{0.48\columnwidth}
\vspace{-2.5em}
\begin{minipage}{0.45\columnwidth}
\begin{algorithm}[H]
\caption{Global Recurrent Step $\mathcal{R}_W$}
\label{alg:global_step}
\begin{algorithmic}[1]
\State Initialize $g_H \gets h_H^{(r-1)}, g_L \gets h_L^{(r-1)}$
\For{$t_H = 1$ to $T_H$}
    \State $g_H \gets \mathcal{R}_{H,W_H}(g_H, g_L)$
    \For{$t_L = 1$ to $T_L$}
        \State $g_L \gets \mathcal{R}_{L,W_L}(g_H, g_L)$
    \EndFor
\EndFor
\State \textbf{return} $(g_L, g_H)$
\end{algorithmic}
\end{algorithm}
\end{minipage}
\vspace{-1em}
\end{wrapfigure}

This global step is formalized in Algorithm~\ref{alg:global_step} as two nested recurrent updates, $\mathcal{R}_{H,W_H}$ and $\mathcal{R}_{L,W_L}$, which exchange information across the hierarchy at different message-passing frequencies. Specifically, $\mathcal{R}_{H,W_H}$ is applied $T_H$ times on $G_H$, and between high-level updates, $\mathcal{R}_{L,W_L}$ is applied $T_L$ times on $G_L$. Each application of $\mathcal R_W$ therefore maps the previous states $h_H^{(r-1)}$ and $h_L^{(r-1)}$ to updated states $g_H$ and $g_L$ for the two hierarchy levels. We now define $\mathcal{R}_{H,W_H}$ and $\mathcal{R}_{L,W_L}$ as follows:
    \begin{equation*}
        \mathcal{R}_{H, W_H}(g_H, g_L) := g_{H} 
        + \mathrm{MP}_{W_H}\!\left(
            \hat{g}_{H}
            + \mathrm{Pool}\!\left(
                \hat{g}_L
            \right)
            ,\,e_H
        \right),
    \end{equation*}
    \begin{equation*}
        \mathcal{R}_{L,W_L}\!\left(
            g_H,\, g_L
        \right)
        :=
        g_L
        + \mathrm{MP}_{W_L'}\!\left(
            f_L
            + \hat{g}_L
            + W_{L}''\, \mathrm{Unpool}\big(g_H\big)
            ,\,e_L
        \right),
    \end{equation*}
where $W_L = (W_L', W_L'')$ denotes the learnable weights of the low-level recurrence. 
We use $\hat{g}$ to denote the normalized state $\hat{g} = \mathrm{RMSNorm}(g)$, following \cite{zhang2019root}, which helps maintain stable feature magnitudes, see Appendix~\ref{sec:RMSNorm} for details. A diagram for the nested recurrent loops in $\mathcal{R_{W}}$ is reported in Figure ~\ref{fig:global_recurrent_step}.
The coarse state enters the low-level update additively, through the projection $W''_L$, so the low-level state is never replaced by or concatenated with the coarse one, and node-level information on $G_L$ is preserved. For $W''_L = 0$, GHR reduces to a flat recurrent MPNN with the same number of low-level steps, so the hierarchy does not restrict the model.
If the coarsening is uninformative, the model can in principle learn to disregard it (Appendix~\ref{pooling-examples}).

The framework is \emph{agnostic} to the choice of the message-passing functions $\mathrm{MP}_{W'_L}$ and $\mathrm{MP}_{W_H}$, any operator of the form \eqref{eq:MP} can be used. Applications tested in Section~\ref{experiments} include GIN~\citep{xu2018how}, GINE \citep{Hu*2020Strategies}, GCN \citep{kipf2017semisupervised} and GatedGCN \citep{bresson2018residual}. 

\paragraph{Readout and Training.}
At each recurrent step $r$, $\mathcal{R}_W$ produces the states $h_L^{(r)}$ and $h_H^{(r)}$, and a task-specific readout maps the low-level state to a prediction. For node-level tasks this is applied node-wise, $\hat{y}_i^{(r)} = W h_{L,i}^{(r)}$. GHR produces a prediction $\hat{y}^{(r)}$ after every global step, and training minimizes the sum of the per-step losses, $\mathcal{L}_{\textnormal{total}} = \sum_{r=1}^{R} \mathcal{L}(\hat{y}^{(r)}, y)$, so that intermediate steps are supervised and gradients reach early iterations of the unrolled computation~\citep{pascanu2013on}. At inference, only the final prediction $\hat{y}^{(R)}$ is used. Readouts for other task types, and the full training and inference procedure, are detailed in Appendix~\ref{app:training}.

\paragraph{Computational Cost.}
\label{sec:cost}
Each update in GHR is a single message-passing layer, whose cost is linear in the size of the graph it runs on. The low-level update acts on $G_L$ and the high-level update on the pooled graph $G_H$, which is never larger in size since $\mathrm{Pool}$ is a quotient map. A global recurrent step performs $T_H$ high-level and $T_H T_L$ low-level updates, and the forward pass applies it $R$ times, so the total cost is
$O\left( R \cdot T_H \cdot T_L \left( |V_L| + |E_L| \right) \right)$
linear in the size of the input graph at fixed depth. Empirical wall-clock and memory measurements are reported in Appendix~\ref{VRAM}.

\begin{figure}
  \begin{center}
  \includegraphics[width=1\linewidth]{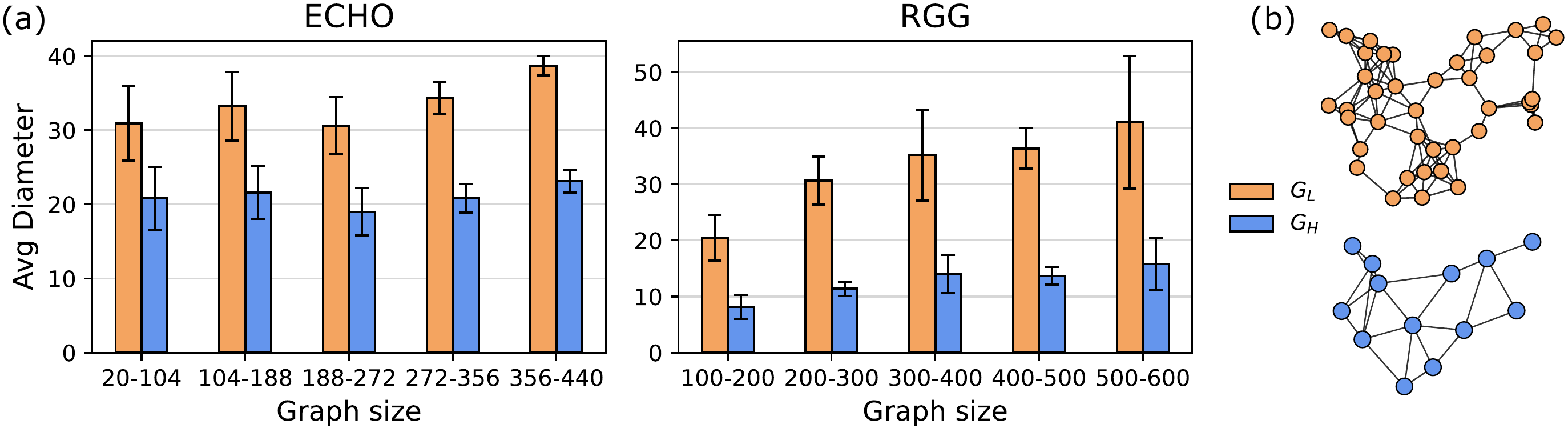}
  \end{center}
  \caption{\textbf{Diameter of graphs and their pooled abstraction.} Comparison of graph diameter between the low-level graph $G_L$ and its high-level abstraction $G_H = \mathrm{Pool}(G_L)$ obtained via graclus. (a) Average diameter of $G_L$ and $G_H$ across the ECHO benchmark~\citep{miglior2025can} and the RGG dataset of Section~\ref{sec:Ablations}, binned by graph size. $G_H$ is obtained with one graclus iteration for ECHO and three for RGG. Error bars show the standard deviation. (b) Visualization of a sample RGG graph $G_L$ (top) and its pooled abstraction $G_H$ (bottom).}
  \label{fig:fig_rgg_diameter}
\vspace{-8pt}
\end{figure}

\subsection{Contraction of distances on $G_{H}$}
\label{sec:contraction}

 By reducing the required propagation steps, the hierarchy in GHR serves to mitigate \emph{topological} over-squashing, which arises from structural bottlenecks in the graph and can depend exponentially on the shortest-path distance between nodes~\citep{diGiovanniOversquashing}.
 Let $d_L : V_L \times V_L \to \mathbb{R}$ denote the shortest-path distance between nodes in $G_L$, and define $d_H$ analogously on $G_H$. 
 The pooling maps we adopt are 1-Lipschitz with respect to these distances: $d_H(\mathrm{Pool}(u), \mathrm{Pool}(v)) \leq d_L(u, v) $ for all $ u, v \in V_L$.
 This property holds if $\mathrm{Pool}$ is a graph quotient map.
 i.e., when $V_H$ is a partition of $V_L$ and $E_H$ contains an edge between two nodes in $V_{H}$ whenever any pair of their original nodes in $V_{L}$ was connected \citep{bader2013graph}. 
The inequality follows from the fact that any path in $G_L$ projects to a path of equal or shorter length in $G_H$. Hence, the high-level graph has equal or smaller diameter ($\mathrm{max}(d_{H})$) than the low-level graph, as shown in Figure~\ref{fig:fig_rgg_diameter}.
The contraction of distances on $G_H$ reduces the number of message-passing steps required to propagate information between two nodes. In a flat MPNN, each MP layer expands the receptive field by one hop, mirroring the round-based algorithms in the LOCAL model of distributed computing~\citep{linial1992locality}. 
Hence information from $u$ reaches $v$ only after at least $d_L(u, v)$ MP steps. In GHR, through the hierarchy, information from $u$ is pooled, propagated over $G_H$ and unpooled onto $v$, so that the number of high-level MP steps required depends on $d_H$ rather than on $d_L$. This increases the receptive field per step with respect to message passing on the original graph.

\section{Experiments}
\label{experiments}

We evaluate GHR across diverse domains, comparing it against linear-complexity baselines (as GHR) and quadratic-complexity graph transformers.
In Section~\ref{sec:Ablations} we first isolate the contribution of GHR's architectural components in a controlled OOR setting. Then in Section~\ref{sec:benchmarks} we evaluate the framework on established long-range benchmarks, where it achieves state-of-the-art or competitive performance, and assess its consistency across message-passing backbones.\\
For each benchmark, we use the official data splits and training protocols. All results are reported as the mean and standard deviation over three seeds.
Baseline values are taken from the original benchmarks unless stated otherwise.

\subsection{Controlled Study of Out-of-Range Generalization}
\label{sec:Ablations}

We study OOR generalization in controlled settings where we compute a \textbf{range-stratified MAE} to evaluate model's performance across different interaction ranges: predicting single-source shortest-path (SSSP) distances from a designated source node on unweighted Random Geometric Graphs (RGG). We remark that in this experiment, no spatial coordinates are provided to the model. Training and test graphs share the same graph size distribution, and differ only in the distances of target nodes. This separates the interaction range from graph size, unlike settings that extrapolate by increasing the graph sizes. We first show that this setting isolates an out-of-range effect rather than under-reaching \citep{barcelo2020logical} issue. We then provide architectural ablations on GHR to highlight the role of recurrence and hierarchy in OOR generalization.

\paragraph{Small-distance regime OOR generalization}


\begin{figure}[h!]
\begin{center}
  \centering
  \includegraphics[width=1\linewidth]{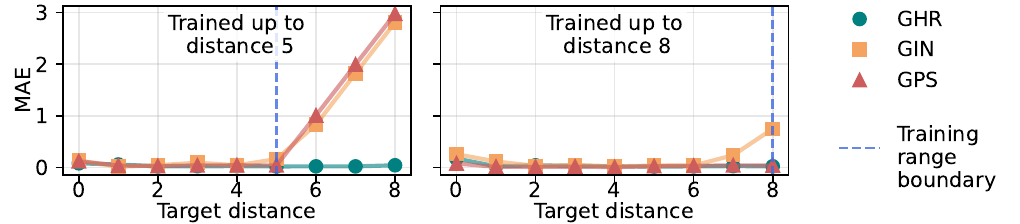}
  \end{center}
  \caption{\textbf{OOR generalization on RGG-SSSP, small-distance regime.} MAE stratified by target distance for GHR-GIN, a deep GIN baseline, and a GPS-style baseline.
    Models are trained with maximum distance 5 (left) or 8 (right), and evaluated on test graphs with distances up to 8.}
  \label{fig:GHR_vs_GPS}
\end{figure}


To verify that errors beyond the training range reflect an out-of-range effect rather than under-reaching \citep{barcelo2020logical,errica2025adaptive}, we run a small-distance version of SSSP on RGGs with $40$--$60$ nodes, comparing GHR with two $10$-layer fixed-depth models, a GIN stack and a GPS model with global attention.
When trained on distances up to $5$ (hops) and tested up to $8$, both baselines degrade approximately linearly beyond the training range (MAE $\approx 3$ at distance $8$), while GHR remains accurate. When trained up to $8$ hops, all three are accurate (Figure~\ref{fig:GHR_vs_GPS}).  We attribute the fist failure to the interaction distances absent from training, not to insufficient receptive field due to model depth. More details are in Appendix~\ref{sec:rgg_oor_small}.

\paragraph{Ablation Study.}

Here models are trained on RGGs whose distances up to $21$ hops and evaluated on graphs containing distances up to $40$. 
GHR combines two mechanisms, weight-shared recurrence and hierarchical abstraction. We compare five variants that isolate their contributions, at a fixed low-level computational budget.
Recurrence appears in two ways, whether the layers share parameters across steps (``Recursive"), and whether the layer(s) are iterated under the Global Recurrent step (``+GR", Eq.~\ref{eq:rs}), carrying the hidden state and an intermediate prediction across $R$. The hierarchy enters with the high-level stream on $G_{H}$, which combined with GR defines GHR. DeepGHR has the same hierarchical coupling but is not recurrent.
All variants use GIN as the backbone and perform $30$ low-level message-passing steps; GHR uses $R = 2$, $T_H = 3$, $T_L = 5$. 
We match the low-level budget so that the differences between GHR and the recurrent variants can be attributed to the coarse stream.
We also report on unrolled versions, in which the low-level recurrence is extended to $42$ at inference only. Additional details on models and configurations are given in Appendix~\ref{sec:sssp_RGG_details}.

Figure~\ref{fig:ablations-stratified} reports MAE stratified by target distance. Dashed lines mark the training boundary ($21$ hops) and the low-level reach ($30$ hops). Fixed-depth variants degrade immediately past the training range despite their larger reach.
Recurrent variants extrapolate the learned update better beyond the training range, with approximately linear error growth only past the reach boundary. GHR has the lowest error: at $30$ hops, its MAE remains below $1.0$, with a lower slope beyond this threshold suggesting a hierarchy contribution.
\rev{Recurrent models are also tested with additional recurrent steps at inference time ("unrolled"). RecurrentGIN improves OOR but loses accuracy IR, whereas its "+GR" variant is more stable in range but has limited improvement OOR. Unrolled GHR maintains IR accuracy while obtaining the lowest error OOR. 
We therefore observe that in this SSSP task, when recurrent models are further unrolled beyond the trained receptive field, global recurrence aids stability while the hierarchical MP stream reduces error in the OOR regime.}

\begin{figure}[h]
  \begin{center}
  \includegraphics[width=0.99\linewidth]{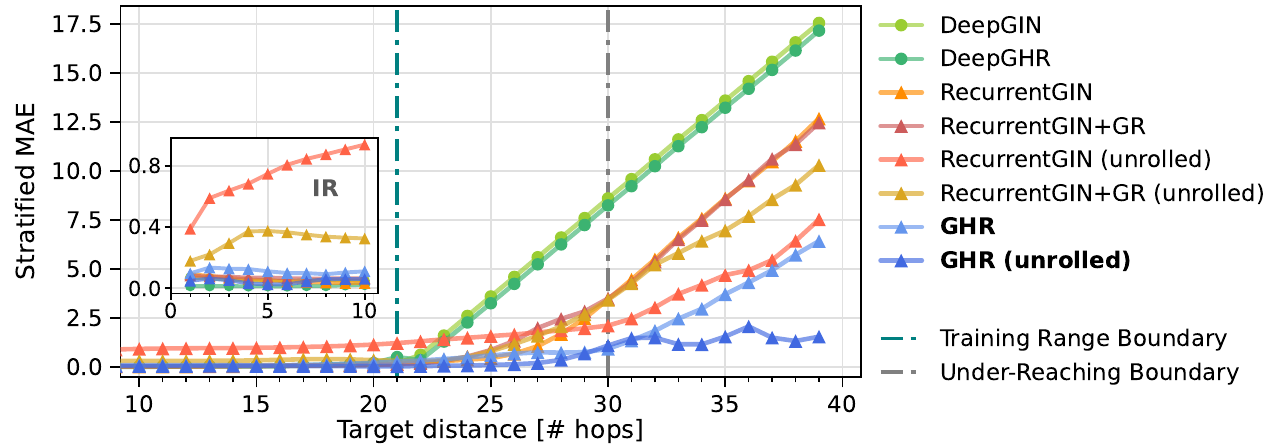}
  \end{center}
    \caption{\textbf{Architectural ablations of GHR on RGG SSSP.} MAE stratified by target distance for the variants described in Section~\ref{sec:Ablations}. Models are trained on distances up to $21$ hops and evaluated on distances up to $40$, with a budget of $30$ message-passing operations both in training and test. \rev{Recurrent "unrolled" models are evaluated extending recursion ($42$ steps) at inference time.}}
  \label{fig:ablations-stratified}
\end{figure}

\vspace{-8pt}
\subsection{Benchmarks}
\label{sec:benchmarks}

We study GHR across benchmarks with three widely used message-passing backbones: GCN~\citep{kipf2017semisupervised}, GIN~\citep{xu2018how} or its edge-aware variant GINE~\citep{Hu*2020Strategies},
and GatedGCN~\citep{bresson2018residual}. We use the published version of these operators as $\mathrm{MP}_W$ in Eq.~\ref{eq:MP} (implementation details in Appendix~\ref{sec:additional_arch}), which allows a direct comparison between each GHR variant and its flat counterpart wherever the benchmark reports the stand-alone backbone.
The recurrence depth $(R, T_H, T_L)$ is selected per task and shared across backbones, so that performance differences are attributable to the message-passing operator alone. Implementation details and hardware specifications are in Appendix \ref{supp_results}. 

Results on \textbf{LRGB} benchmark are reported in Appendix~\ref{sec:supplementary_results}, where we improve performance on each MP backbone in both studied tasks.
Following Section~\ref{background}, we use graclus~\citep{dhillon2007weighted} for ECHO, LRGB, and RGG. For graphs with regular structure, the coarsening can exploit the underlying symmetries: on the LRIM lattices, we replace graclus with a fixed $2 \times 2$ block partition. Appendix~\ref{app:pooling-ablation} evaluates alternative coarsenings, including random partitions and excessive compression, and characterizes the cases in which the coarsening is not suited to the task, with an ablation study. Details on clustering and feature aggregation in Appendix~\ref{pooling-examples}.

\paragraph{ECHO:}
We test GHR on the ECHO benchmark~\citep{miglior2025can} which covers two regimes. ECHO-Synth consists of algorithmic tasks (single-source shortest paths, diameter, and eccentricity) on synthetic graphs designed to stress models with \emph{topological} bottlenecks. ECHO-Chem consists of regression of DFT-computed atomic charges and molecular energies, where long-range interactions arise in \emph{real-world} molecules. 
ECHO-Synth tests the algorithmic alignment of the model, while ECHO-Chem tests its expressiveness on real-world continuous interactions.
Each backbone improves when placed inside GHR. On all ECHO tasks, both GHR-GCN and GHR-GINE reduce the error of their stand-alone counterparts (Table~\ref{tab:combined-results}). GHR-GINE further achieves the best reported result on all three ECHO-Synth tasks and on ECHO-Charge, and is second only to GPS on ECHO-Energy, which has quadratic complexity in the number of nodes. This indicates that hierarchical recurrence enables effective long-range propagation even under strong topological bottlenecks, consistent with the diameter reduction of the pooled graph reported in Figure~\ref{fig:fig_rgg_diameter}.

\begin{table}[ht]
\vspace{-6pt}
\caption{\textbf{ECHO benchmark.} Performance comparison (Mean Absolute Error - MAE $\downarrow$) across synthetic and chemical tasks. Baseline values are reported directly from \citep{miglior2025can}. Charge results are reported in units of $10^{-3}$.  \textcolor[HTML]{009e74}{\textbf{Best}}, \textcolor[HTML]{d55e00}{\textbf{second}}, and \textbf{third} per column.}
\label{tab:combined-results}
\begin{center}
\footnotesize
\setlength{\tabcolsep}{4pt}
\setlength{\aboverulesep}{0pt}
\setlength{\belowrulesep}{0pt}
\renewcommand{\arraystretch}{1.05}
\begin{tabular}{l ccc cc}
\toprule
& \multicolumn{3}{c}{\textbf{ECHO-Synth}} & \multicolumn{2}{c}{\textbf{ECHO-Chem}} \\
\cmidrule(r){2-4} \cmidrule(l){5-6}
\textbf{Model} & \textbf{sssp} $\downarrow$ & \textbf{ecc} $\downarrow$ & \textbf{diam} $\downarrow$ & \textbf{Charge} $\downarrow$ & \textbf{Energy} $\downarrow$ \\
\midrule
GHR-GINE (Ours) & \textcolor[HTML]{009e74}{$\mathbf{0.035_{\pm 0.004}}$} & \textcolor[HTML]{009e74}{$\mathbf{3.456_{\pm 0.041}}$} & \textcolor[HTML]{009e74}{$\mathbf{0.749_{\pm 0.023}}$} & \textcolor[HTML]{009e74}{$\mathbf{5.400_{\pm 0.035}}$} & \textcolor[HTML]{d55e00}{$\mathbf{7.152_{\pm 0.263}}$} \\
GHR-GCN (Ours)  & $0.759_{\pm 0.032}$ & $4.815_{\pm 0.242}$ & $1.789_{\pm 0.138}$ & $7.936_{\pm 0.018}$ & $\mathbf{11.178_{\pm 0.381}}$ \\
\midrule
A-DGN    & $1.176_{\pm 0.140}$ & $4.981_{\pm 0.037}$ & $1.151_{\pm 0.038}$ & $6.543_{\pm 0.146}$ & $12.486_{\pm 1.621}$ \\
DRew     & $1.279_{\pm 0.011}$ & \textcolor[HTML]{d55e00}{$\mathbf{4.651_{\pm 0.020}}$} & $1.243_{\pm 0.047}$ & $9.086_{\pm 0.473}$ & $11.325_{\pm 2.394}$ \\
GCN      & $2.102_{\pm 0.094}$ & $5.233_{\pm 0.034}$ & $3.832_{\pm 0.262}$ & $8.421_{\pm 0.512}$ & $28.112_{\pm 1.239}$ \\
GCNII    & $2.128_{\pm 0.429}$ & $5.241_{\pm 0.030}$ & $2.005_{\pm 0.093}$ & $8.829_{\pm 0.021}$ & $13.235_{\pm 2.630}$ \\
GIN/GINE & $2.234_{\pm 0.271}$ & $4.869_{\pm 0.092}$ & $1.630_{\pm 0.161}$ & $7.176_{\pm 0.371}$ & $23.558_{\pm 7.568}$ \\
GPS      & $\mathbf{0.472_{\pm 0.050}}$ & $\mathbf{4.758_{\pm 0.021}}$ & $2.160_{\pm 0.098}$ & $\mathbf{6.182_{\pm 0.219}}$ & \textcolor[HTML]{009e74}{$\mathbf{5.257_{\pm 0.842}}$} \\
GraphCON & $5.734_{\pm 0.011}$ & $5.474_{\pm 0.001}$ & $2.969_{\pm 0.189}$ & $19.629_{\pm 0.195}$ & $14.295_{\pm 0.807}$ \\
GRIT     & \textcolor[HTML]{d55e00}{$\mathbf{0.121_{\pm 0.013}}$} & $5.091_{\pm 0.158}$ & \textcolor[HTML]{d55e00}{$\mathbf{1.014_{\pm 0.046}}$} & $7.134_{\pm 6.090}$ & $25.508_{\pm 2.507}$ \\
PH-DGN   & $1.323_{\pm 0.485}$ & $5.068_{\pm 0.126}$ & $1.627_{\pm 0.398}$ & $7.915_{\pm 0.269}$ & $16.080_{\pm 1.123}$ \\
SWAN     & $0.896_{\pm 0.232}$ & $4.840_{\pm 0.045}$ & $\mathbf{1.121_{\pm 0.070}}$ & \textcolor[HTML]{d55e00}{$\mathbf{6.109_{\pm 0.103}}$} & $12.629_{\pm 1.157}$ \\
\bottomrule
\end{tabular}
\end{center}
\end{table}

In ECHO-Synth, training and test graphs share the same distribution of diameters and distances, so the benchmark measures in-range generalization only. To evaluate out-of-range generalization on these tasks, each dataset is masked on the target value. Models are trained on instances below a task-specific threshold and evaluated separately within that range (IR) and outside it (OOR) with, additional details in Appendix~\ref{app:echo-oor}. All models use the configuration selected for the corresponding task in the original benchmark~\citep{miglior2025can}, and none performs additional iterations at inference. Table~\ref{tab:echo-synth-oor} reports the results and the task-specific range thresholds. GHR obtains the lowest error on all three tasks in both regimes, and the margin is consistently larger out-of-range.

\begin{table}[ht]
\caption{\textbf{Performance on OOR generalization on ECHO-Synth.} GHR-GINE was the strongest model on ECHO, here it is tested by trained on instances whose target is in the in-range interval and evaluated both on that interval (IR) and beyond it (OOR). Selected baselines are: a recurrent model (ADGN), a re-wiring model (DReW) and two transformers (GRIT, GPS). All models use the configuration from the corresponding task in \citep{miglior2025can}, with no additional iterations at inference. Metrics represent MAE ($\downarrow$). Best results are in bold.}
\label{tab:echo-synth-oor}
\begin{center}
\footnotesize
\setlength{\tabcolsep}{4pt}
\setlength{\aboverulesep}{0pt}
\setlength{\belowrulesep}{0pt}
\renewcommand{\arraystretch}{1.05}
\begin{tabular}{l cc cc cc}
\toprule
& \multicolumn{2}{c}{\textbf{sssp}} & \multicolumn{2}{c}{\textbf{ecc}} & \multicolumn{2}{c}{\textbf{diam}} \\
\cmidrule(lr){2-3} \cmidrule(lr){4-5} \cmidrule(lr){6-7}
\textbf{Model} & \textbf{IR} ($1$--$21$) & \textbf{OOR} ($22$--$40$) & \textbf{IR} ($9$--$30$) & \textbf{OOR} ($31$--$40$) & \textbf{IR} ($17$--$30$) & \textbf{OOR} ($31$--$40$) \\
\midrule
A-DGN & $0.181_{\pm 0.057}$ & $5.182_{\pm 0.058}$ & $3.751_{\pm 0.012}$ & $10.267_{\pm 0.298}$ & $1.179_{\pm 0.002}$ & $2.015_{\pm 0.048}$ \\
DRew  & $0.579_{\pm 0.017}$ & $8.686_{\pm 0.029}$ & $3.604_{\pm 0.043}$ & $10.363_{\pm 0.151}$ & $1.693_{\pm 0.139}$ & $4.143_{\pm 0.136}$ \\
GPS   & $0.340_{\pm 0.115}$ & $7.101_{\pm 0.461}$ & $3.889_{\pm 0.017}$ & $10.868_{\pm 0.306}$ & $3.916_{\pm 0.231}$ & $8.057_{\pm 0.331}$ \\
GRIT  & $0.464_{\pm 0.170}$ & $6.803_{\pm 1.234}$ & $3.718_{\pm 0.029}$ & $10.111_{\pm 0.092}$ & $3.815_{\pm 0.526}$ & $7.338_{\pm 1.169}$ \\
\midrule
\textbf{GHR-GINE}  & $\mathbf{0.052_{\pm 0.041}}$ & $\mathbf{0.226_{\pm 0.051}}$ & $\mathbf{3.086_{\pm 0.054}}$ & $\mathbf{7.403_{\pm 0.214}}$ & $\mathbf{1.112_{\pm 0.044}}$ & $\mathbf{1.751_{\pm 0.100}}$ \\

\bottomrule
\end{tabular}
\end{center}
\end{table}
\vspace{-4pt}

\paragraph{LRIM:}

The Long-Range Ising Model (LRIM) benchmark \citep{mathys2026lrim} evaluates dependencies through an Ising Hamiltonian with power-law couplings. The task is a node regression of local energy changes for spin configurations on a 2D lattice. Each target is an all-pairs sum over every pair of spins, so the task requires compressing information from the entire lattice into each node representation, a \emph{computational} bottleneck that grows with the lattice size. We report the results on \textbf{LRIM-hard} in Table~\ref{tab:lrim-16-hard}, where distant spins contribute to every target. Additional details are in Appendix~\ref{supp_results}.
Both backbones improve all the baselines when placed inside GHR, on both lattice sizes, and GHR-GatedGCN attains the best reported result on each. All these results are obtained without feeding any positional encoding to the models. 
We remark that models are implemented with only $9$ and $18$ low-level message-passing steps on \emph{16-hard} and \emph{32-hard}, respectively (\ref{app:hyperparameters}). A flat MPNN with the same number of steps sees at most a $9$ or $18$ hop neighbourhood, and even the oracle of Ref. \citep{mathys2026lrim} (Figure 2), which computes the exact target restricted to a neighbourhood, has a higher error than GHR. Since each high-level step spans multiple low-level hops, the coarse stream extends the receptive field beyond the number of low-level consistently with Section~\ref{sec:contraction}.

\begin{table}[ht]
\vspace{-6pt}
\caption{\textbf{Baseline Performance on LRIM-16-hard and LRIM-32-hard.} Values represent LogMSE. Baseline values are reported from the original benchmark literature \citep{mathys2026lrim}. \textcolor[HTML]{009e74}{\textbf{Best}}, \textcolor[HTML]{d55e00}{\textbf{second}}, and \textbf{third} per column.}
\label{tab:lrim-16-hard}
\begin{center}
\footnotesize
\setlength{\tabcolsep}{6pt}
\setlength{\aboverulesep}{0pt}
\setlength{\belowrulesep}{0pt}
\renewcommand{\arraystretch}{1.05}
\begin{tabular}{lcccc}
\toprule
\textbf{Model} & \textbf{LRIM-16-hard} $\downarrow$ & \textbf{LRIM-32-hard} $\downarrow$\\
\midrule
GHR-GIN (Ours) & \textcolor[HTML]{d55e00}{$\mathbf{-4.378_{\pm 0.005}}$} & \textcolor[HTML]{d55e00}{$\mathbf{-4.220_{\pm 0.011}}$} \\
GHR-GatedGCN (Ours) & \textcolor[HTML]{009e74}{$\mathbf{-4.701_{\pm 0.031}}$} & \textcolor[HTML]{009e74}{$\mathbf{-4.852_{\pm 0.013}}$} \\
\cmidrule(r){2-3}
GIN & $-2.533_{\pm 0.313}$ & $-2.249_{\pm 0.135}$ \\
GatedGCN & $-3.844_{\pm 0.055}$ & $-4.087_{\pm 0.238}$ \\
GatedGCN-VN$_G$ & $-4.068_{\pm 0.131}$ & $-3.243_{\pm 0.170}$ \\
GPS-Base & $-4.211_{\pm 0.155}$ & $-4.044_{\pm 0.122}$ \\
GPS-RWSE & $-4.011_{\pm 0.129}$ & $\mathbf{-4.134_{\pm 0.075}}$ \\
GPS-LapPE & $\mathbf{-4.334_{\pm 0.065}}$ & $-4.032_{\pm 0.092}$ \\
\bottomrule
\end{tabular}
\end{center}
\end{table}

Following Ref. \citep{mathys2026lrim}, we also evaluate out-of-domain transfer from LRIM-16-hard to larger lattices (Table~\ref{tab:lrim-transfer-short}). This is not out-of-range according to our definition. A larger lattice increases the interaction range, but also changes size and target distribution. Still, both GHR implementation improve every baseline at all sizes, with the graph transformers running out of memory on the largest lattice even at inference.

\begin{table}[ht]
\vspace{-6pt}
\caption{\textbf{Transfer performance for models trained on LRIM-16-hard to OOD lattice dimensions.} Values represent LogMSE. Full table with all baselines is reported in Appendix~\ref{LRIM-extended}.\textcolor[HTML]{009e74}{\textbf{Best}}, \textcolor[HTML]{d55e00}{\textbf{second}}, and \textbf{third} per column.}
\label{tab:lrim-transfer-short}
\begin{center}
\footnotesize
\setlength{\tabcolsep}{6pt}
\setlength{\aboverulesep}{0pt}
\setlength{\belowrulesep}{0pt}
\renewcommand{\arraystretch}{1.05}
\begin{tabular}{lcccc}
\toprule
\textbf{Model} & \textbf{32-hard} $\downarrow$ & \textbf{64-hard} $\downarrow$ & \textbf{128-hard} $\downarrow$ & \textbf{256-hard} $\downarrow$ \\
\midrule
GHR-GIN (Ours) & \textcolor[HTML]{d55e00}{$\mathbf{-1.085_{\pm 0.021}}$} & \textcolor[HTML]{009e74}{$\mathbf{-0.843_{\pm 0.030}}$} & \textcolor[HTML]{d55e00}{$\mathbf{-0.770_{\pm 0.011}}$} & \textcolor[HTML]{d55e00}{$\mathbf{-1.066_{\pm 0.033}}$} \\
GHR-GatedGCN (Ours) & \textcolor[HTML]{009e74}{$\mathbf{-1.086_{\pm 0.014}}$} & \textcolor[HTML]{d55e00}{$\mathbf{-0.825_{\pm 0.023}}$} & \textcolor[HTML]{009e74}{$\mathbf{-0.771_{\pm 0.001}}$} & \textcolor[HTML]{009e74}{$\mathbf{-1.073_{\pm 0.003}}$} \\
\midrule
GIN & $-1.043_{\pm 0.051}$ & $-0.774_{\pm 0.042}$ & $-0.703_{\pm 0.041}$ & $-0.903_{\pm 0.047}$ \\
GatedGCN & $-1.050_{\pm 0.004}$ & $-0.781_{\pm 0.002}$ & $-0.708_{\pm 0.003}$ & $\mathbf{-0.952_{\pm 0.005}}$ \\
GPS-Base & $\mathbf{-1.057_{\pm 0.000}}$ & $\mathbf{-0.790_{\pm 0.001}}$ & $\mathbf{-0.719_{\pm 0.001}}$ & OOM \\
\bottomrule
\end{tabular}
\end{center}
\end{table}

\vspace{-10pt}


\section{Conclusion}

We identified a fundamental limitation of current graph learning architectures: strong performance on long-range dependencies does not guarantee extrapolation beyond distances observed during training. To expose this gap, we introduced \emph{out-of-range generalization} as a new evaluation regime. 
To address this challenge, we introduced Graph Hierarchical Recurrence (GHR), which couples recurrent computation with hierarchical graph abstractions. By jointly operating on the input graph and its high-level representations, GHR enables efficient information propagation across large distances while preserving the underlying topology. This design provides a principled mechanism to scale recurrent depth without increasing model size.
Across diverse long-range benchmarks, GHR delivers strong performance and consistently improves every evaluated backbone.
More broadly, these results point to a direction complementary to architecture scaling: designing mechanisms that explicitly support out-of-range generalization.
Our findings position the combination of recurrence and hierarchical representations as a promising basis for scalable graph models that remain effective beyond their training propagation regime.
A natural next step is to extend GHR beyond two hierarchical levels, potentially reducing the number of recurrent iterations needed to bridge long distances.
The interplay between hierarchy and recurrence also raises broader theoretical questions about expressivity and receptive fields. Weight sharing constrains every recurrent step to apply the same update, whereas the coarse stream enlarges the distance covered at each step. Characterizing the functions representable by such an inductive bias, and how the coarse graph control its effective propagation range, defines an important direction for future work.

\newpage

\begin{ack}
AG and DB acknowledge funding from EU-EIC EMERGE (Grant No. 101070918). MP, BL, and SB acknowledge funding from Ministero delle Imprese e del Made in Italy (IPCEI Cloud DM 27 giugno 2022 – IPCEI-CL-0000007) and European Union (Next Generation EU), as well as from EU Horizon projects TANGO (No. 101120763) and ELIAS (No. 101120237).  

\end{ack}

\bibliographystyle{plain}

\bibliography{references}


\appendix

\section{Additional details on the architecture}
\label{sec:additional_arch}

In this section additional information on the architectural elements of GHR, the message-passing backbones adopted and training and inference details are provided.

\subsection{Input Encoding}
\label{app:input-encoding}
Node and edge features are projected into the $m$-dimensional latent space of $\mathcal{R}_W$ via learnable linear maps, as shown in Algorithm~\ref{alg:input-encoding}: the low-level node features are encoded as $f_{L,i} = W_n \mathbf{x}_{L,i}$ for $i \in V_L$, and the edge features of $G_L$ and $G_H$ by $W_{e_L}$ and $W_{e_H}$, respectively, retaining the notation $\mathbf{e}_L$ and $\mathbf{e}_H$ for the encoded edge features. The projections are computed once per forward pass, and their outputs are injected unchanged at each iteration of the global recurrent step.

\begin{algorithm}[h]
\caption{Input Encoding}\label{alg:input-encoding}
\begin{algorithmic}[1]
\State $\mathbf{e}_{H,\alpha} \gets W_{e_H}\, \mathbf{e}_{H,\alpha}$ \hfill $\forall \alpha \in E_H$
\State $\mathbf{e}_{L,\beta} \gets W_{e_L}\, \mathbf{e}_{L,\beta}$ \hfill $\forall \beta \in E_L$
\State $f_{L,i} \gets W_n\, \mathbf{x}_{L,i}$ \hfill $\forall i \in V_L$
\end{algorithmic}
\end{algorithm}

The initial hidden states are obtained from two learnable vectors $h_L^{\mathrm{init}}, h_H^{\mathrm{init}} \in \mathbb{R}^m$, shared across nodes and broadcast to every node of $G_L$ and $G_H$, respectively. Since they are parameters of the model, the initial states are identical for all graphs at inference time.


\subsection{Message-Passing Instantiations}
\label{app:mp}

The framework is agnostic to the choice of message-passing operator: any function of the form \eqref{eq:MP} can be used as $\mathrm{MP}_{W_L}$ and $\mathrm{MP}_{W_H}$, and the two levels may use the same operator or different ones. In our experiments both levels use the same operator, selected among three widely adopted architectures. In all cases we use the published formulation of the operator, embedded in the recurrent update of Algorithm~\ref{alg:global_step}, which additionally applies a residual connection and normalizes the state before aggregation (Appendix~\ref{sec:RMSNorm}).

\textbf{GCN.} We use the convolution of Ref. \citep{kipf2017semisupervised}, which aggregates neighbouring states with symmetric degree normalization. As it does not process edge features, $\mathbf{e}$ is unused when this operator is selected.

\textbf{GIN and GINE.} We use GIN \citep{xu2018how} on graphs without edge features and its edge-aware variant GINE \citep{Hu*2020Strategies} otherwise. We treat the update MLP as a hyperparameter, selecting between a SiLU-activated and a SwiGLU-activated variant \citep{shazeer2020glu}. 
The aggregation is unchanged in either case. 
The selected choice for each task is reported in Appendix~\ref{app:hyperparameters}.
\textbf{GatedGCN.} We use the residual gated convolution of Ref. \citep{bresson2018residual}.

\begin{figure}[h]
  \centering
  \includegraphics[width=1\linewidth]{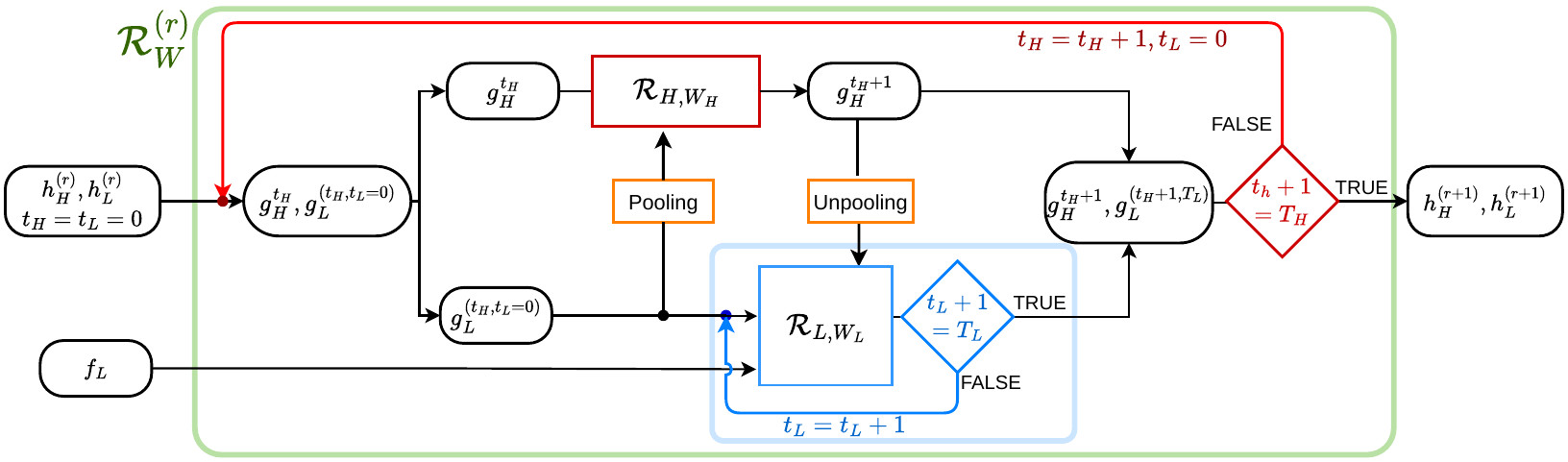}
  \caption{\textbf{The Global Recurrent Step $\mathcal{R}_W$.} Visualization of Algorithm~\ref{alg:global_step}, with the low-level update $\mathcal{R}_{L,W_L}$ in blue and the high-level update $\mathcal{R}_{H,W_H}$ in red. The states $g_L^{(t_H, t_L)}$ and $g_H^{(t_H)}$ are indexed by their timestamp within the nested recursion. The architecture executes $T_H$ high-level iterations. Within each iteration, the low-level module unrolls for $T_L$ steps. After $T_H$ outer iterations, the states $h_L^{(r+1)}$ and $h_H^{(r+1)}$ are returned for the next global reasoning step $r+1$, as described in Figure \ref{fig:full_architecture}.}
  \label{fig:global_recurrent_step}
\end{figure}

\begin{figure}[h]
  \begin{center}
  \includegraphics[width=1\linewidth]{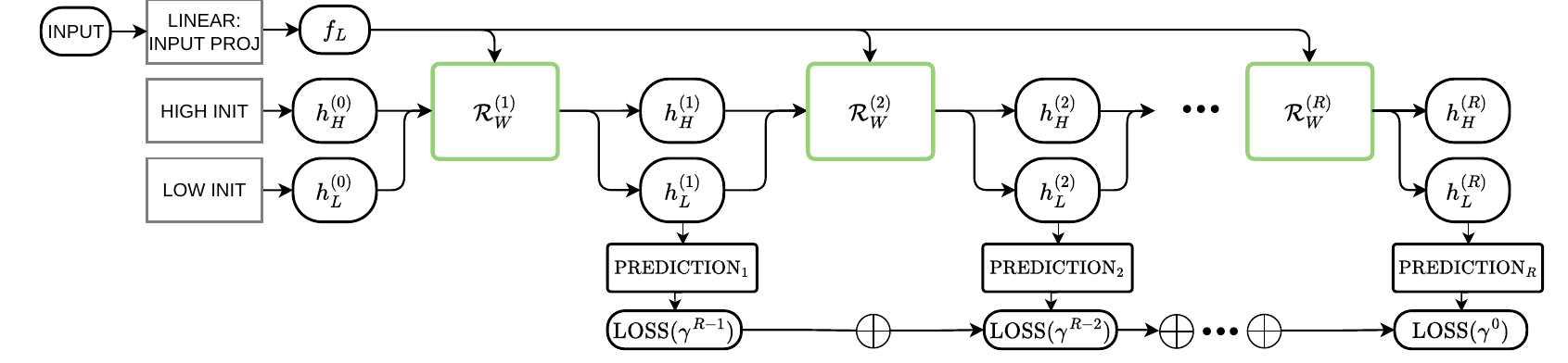}
  \end{center}
  \caption{\textbf{Complete recurrent architecture}. Full recurrent architecture with the global recursion scheme and time-discounted loss computation (complete forward pass) as described in Section \ref{model}. }
  \label{fig:full_architecture}
\end{figure}

\subsection{Architecture Diagrams}

Two diagrams are provided to better represent the global recursion scheme and the nested recursion loops connected via pooling-unpooling.
In Figure \ref{fig:full_architecture} the complete global recurrence scheme with time-discounted loss is shown, while in Figure \ref{fig:global_recurrent_step} a single global recurrent step $\mathcal{R}_{W}$ with its nested recurrent loops is provided.

\subsection{Readouts, Training, and Inference}
\label{app:training}
\paragraph{Task-specific readouts.}
GHR is adapted to the task of interest by selecting an appropriate readout layer applied to the low-level state $h_L^{(r)}$. For node-level tasks the readout is applied node-wise, $\hat{y}_i^{(r)} = W h_{L,i}^{(r)}$. For graph-level tasks, the low-level states are first aggregated into a single graph representation, $z^{(r)} = \mathrm{Readout}\big(\{\!\{ h_{L,i}^{(r)} \}\!\}_{i \in V_L}\big)$, and the prediction is obtained as $\hat{y}^{(r)} = W z^{(r)}$. The model returns $\hat{y}^{(r)}$ together with $(h_L^{(r)}, h_H^{(r)})$, so that recurrence can continue at step $r+1$.

\paragraph{Training.}
The trainable parameters are $\hat W = (W_{e_H}, W_{e_L}, W_n, W, W_{\mathrm{ro}})$, collecting the encoding layer (Section~\ref{app:input-encoding}), the global recurrent step (Section~\ref{section:reasoning_step}), and the task-specific readout. Training uses full backpropagation through time over the $R \times T_H \times T_L$ unrolled steps as described in Section~\ref{section:reasoning_step}. The choice of an intermediate rather than a final-step loss is motivated by the large diameters typical of spatial topologies \citep{barthelemy2011spatial}, which require many iterations for information to propagate. Despite the depth, linear complexity and a compact hidden dimension maintain the backward-pass memory footprint low. see Appendix~\ref{VRAM} for the more detailed study.

\paragraph{Inference.}
The intermediate predictions $\hat y^{(r)}$ for $r < R$ are not needed at test time. The input graph $G_L$ is first pooled to obtain $G_H := \mathrm{Pool}(G_L)$, both graphs are encoded through learnable maps (Section~\ref{app:input-encoding}), $R$ steps of $\mathcal{R}_W$ are applied (Section~\ref{section:reasoning_step}), and the readout yields $\hat y^{(R)}$.

\paragraph{Step-conditioned message passing.}
A hyperparameter controls whether the constant input injected at each low-level update also includes an encoding of the step indices $t_H, t_L)$, allowing the update to count its step within the nested recurrence. Hyperparameter search selected this encoding for LRIM, while not using it across the other benchmarks. When enabled, a learnable linear embedding of $(t_H, t_L)$ is added to the input of $\mathrm{MP}_{W'_L}$:
\begin{equation}
\label{time-informed}
\mathcal{R}_{L,W_L}(g_H, g_L) := g_L + \mathrm{MP}_{W'_L}\!\left(f_L + \hat{g}_L + W''_L\,\mathrm{Unpool}(g_H) + W^{\mathrm{time}}_L(t_H, t_L),\; \mathbf{e}_L\right).
\end{equation}

\paragraph{Discounted intermediate loss.}
The per-step losses of \eqref{section:reasoning_step} can optionally be weighted by a discount factor, $\mathcal{L}_{\mathrm{total}} = \sum_{r=1}^{R} \gamma^{R-r}\, \mathcal{L}(\hat{y}^{(r)}, y)$ with $\gamma \in (0, 1]$, so that later global steps contribute more to the loss~\citep{pascanu2013on}. We included $\gamma$ in the hyperparameter search, and all reported configurations use $\gamma = 1$, i.e.\ equal weighting across steps.

\subsection{RMSNorm Formulation}
\label{sec:RMSNorm}

For an input vector $\mathbf{x} \in \mathbb{R}^d$, Root Mean Square Normalization (RMSNorm) \cite{zhang2019root} computes the normalized output $y_i$ for each feature $i$ as:

\begin{equation}
    y_i = w_i \frac{x_i}{\text{RMS}(\mathbf{x})}, \quad \text{where} \quad \text{RMS}(\mathbf{x}) = \sqrt{\frac{1}{d} \sum_{j=1}^{d} x_j^2}
\end{equation}

and $w_i$ is a learnable scaling parameter specific to the $i$-th feature.

\section{Implementation Details}
\label{supp_results}

This appendix details the experimental setup, hardware, and task-specific architectural implementations used in the main evaluation, followed by extended results.

\subsection{Computational efficiency.}
\label{VRAM}
Section~\ref{sec:cost} derives the asymptotic cost of GHR. In practice the coarse stream adds a fraction of a low-level step, set by the pooling ratio $|V_H|/|V_L|$, and the pooling map is non-parametric and precomputed, so it does not enter the per-step cost. We also measure training-step time and peak memory directly, against a depth-matched feedforward GIN baseline on the ECHO tasks (Table~\ref{tab:profiling}), using \verb|torch.cuda.max_memory_allocated()| on an NVIDIA GeForce RTX 2080 Ti. GHR remains within a small factor of the flat baseline on both axes, and the overhead is determined by the two message-passing streams performed at each step.

We also measure inference latency at matched computational depth. Averaged over the ECHO-Synth and ECHO-Chem tasks, with $30$ propagation steps and hidden dimension $32$ for all models, per-graph forward-pass times are reported in Table~\ref{tab:latency}. GHR is roughly an order of magnitude faster than the transformer baseline, while incurring a modest per-step overhead relative to a flat MPNN.

\begin{table}[ht]
\caption{\textbf{Training-step time and peak memory.} GHR compared against a depth-matched feedforward GIN baseline on the ECHO benchmark. Effective depth is $R \times T_H \times T_L$ for GHR and the corresponding number of layers for GIN. Results are obtained with equal hidden dimension $d=32$.}
\label{tab:profiling}
\begin{center}
\small
\setlength{\tabcolsep}{10pt}
\begin{tabular}{lccccc}
\toprule
\textbf{Task} & \textbf{Depth} & \textbf{GHR (ms)} & \textbf{GIN (ms)} & \textbf{GHR (MB)} & \textbf{GIN (MB)} \\
\midrule
ECHO-SSSP   & 72 & 180.3 & 144.0 & 1158 & 966 \\
ECHO-ECC    & 24 & 95.1  & 58.2  & 1040 & 631 \\
ECHO-Diam   & 16 & 66.4  & 41.2  & 376  & 248 \\
ECHO-Energy & 12 & 49.3  & 30.7  & 203  & 144 \\
ECHO-Charge & 12 & 46.5  & 28.3  & 277  & 205 \\
\bottomrule
\end{tabular}
\end{center}
\end{table}

\begin{table}[ht]
\caption{\textbf{Inference latency at matched depth.} Per-graph forward-pass time, averaged over the ECHO-Synth and ECHO-Chem tasks. All models use $30$ propagation steps (layers or recurrent iterations) and hidden dimension $d=32$.}
\label{tab:latency}
\begin{center}
\small
\setlength{\tabcolsep}{12pt}
\begin{tabular}{lc}
\toprule
\textbf{Model} & \textbf{Time per graph (ms)} $\downarrow$ \\
\midrule
GIN & $0.110$ \\
GHR & $0.168$ \\
GPS & $1.610$ \\
\bottomrule
\end{tabular}
\end{center}
\end{table}

\section{Supplementary Results}
\label{sec:supplementary_results}

\subsection{Out-of-Range Generalization on RGG SSSP: Comparison with Fixed-Depth Architectures}
\label{sec:rgg_oor_small}

This appendix reports a small-distance out-of-range experiment, separate from the main study of Section~\ref{sec:Ablations}, which verifies that the degradation observed beyond the training range is an out-of-range effect rather than under-reaching \citep{barcelo2020logical,errica2025adaptive}.

\textbf{Baselines:} We compare GHR-GINE against two representative fixed-depth architectures: a 10-layer GPS-style model combining local GINE message passing with global multi-head attention via PyG's \texttt{GPSConv}~\citep{rampasek2022GPS}, and a 10-layer deep model stacking GINE layers~\citep{Hu*2020Strategies}. All three models use hidden dimension $32$. We remark that these results characterize the behaviour of the specific feedforward architectures implemented, rather than the full space of Graph Transformer designs.

\textbf{Dataset:} RGGs with 40 to 60 nodes and average node degree fixed to 6 and keeping only the largest connected component, partitioned into 6000 training, 1000 validation, and 1000 test graphs. Each graph has a single designated source node.

\textbf{Training conditions:} All models are trained until convergence under two conditions: the maximum shortest-path distance from the source in the training graphs is capped at 5 hops (left panel of Figure~\ref{fig:GHR_vs_GPS}) or at 8 hops (right panel). Both are evaluated on test graphs containing distances up to 8 hops.

\textbf{Results:} When trained up to distance 5, all three models achieve near-zero MAE within the training range, but the two fixed-depth baselines degrade rapidly beyond it, with MAE growing approximately linearly to $\sim 3$ hops at target distance 8, while GHR maintains accuracy across this small out-of-range region. When the training range is extended to 8 hops, all three models predict accurately over the entire test range. 
We attribute the failure in the first condition to the interaction distances absent from training, not to insufficient receptive field due to model depth.


\subsection{Out-of-Range Generalization on RGG SSSP: Ablations}
\label{sec:sssp_RGG_details}

This appendix provides the experimental setup and the full results for the ablation reported in Section~\ref{sec:Ablations} (Figure~\ref{fig:ablations-stratified}). We remark that in RGG tasks, no spatial coordinates are provided to the model so that the spatial correlation with target can not be exploited by any model. 
All variants share the same hidden dimension of $32$, and same linear readout, output extrapolation alone cannot account for the differences between them.

\textbf{Dataset:} The Single-Source Shortest Path (SSSP) task requires the model to predict the shortest-path distance from a single designated source node to every other node in an unweighted Random Geometric Graph. The graphs are generated with an average node degree of $6$, by choosing the connection radius $r$ such that each node has on average $6$ other nodes within distance $r$ in the embedding space and then keeping the largest connected component of the graph. Training, validation and test splits contain $6000$, $1000$ and $1000$ graphs respectively, drawn from the same distribution of node counts: graph are sampled uniformely to have between 300 and 350 nodes, so that the splits differ only in the shortest-path distances they contain: training and validation graphs are masked to distances of at most $21$ hops, while test graphs contain distances up to $40$.

\textbf{Computational budget and evaluation regimes.} All variants execute $30$ low-level message-passing operations: Deep variants stack $30$ parameter-distinct layers, Recurrent variants unroll one layer $30$ times, in GHR we set $R = 2$, $T_H = 3$, $T_L = 5$ resulting in a total of 30 low-level recursive steps.
GHR-deep has the hierarchical coupling of GHR but removes weight sharing. It alternates high-level updates on $G_H$ with low-level updates on $G_L$ coupled through pooling and unpooling, but with distinct parameters at every step: each of its $30$ low-level layers, $6$ high-level layers, and the corresponding unpooling projections has dedicated weights. We use a single global step ($R = 1$) with $T_H = 6$ and $T_L = 5$, so that GHR-deep performs the same numbers of low- and high-level operations as GHR. Since its update differs at every step, DeepGHR cannot be unrolled beyond its trained depth. 
All the hierarchical variants add $6$ operations on the coarse graph ($\approx 1/4$ the node count), introducing a small compute overhead while matching the base low-level budget to isolate the structural effect of coarse routing. Only GHR (Unrolled) increases iteration count, setting $T_L = 7$ post-training strictly during inference.
This low-level matched budget creates two distinct evaluation intervals past the training boundary. Between the training limit and $30$ hops, flat variants have sufficient receptive field to cover the target distance, thus isolating out-of-range generalization from physical under-reaching. Past $30$ hops, flat baselines are subject to under-reaching issues. In this case, the performance of GHR confirms that the coarse stream $G_H$ influences the prediction by compressing path lengths, enabling propagation across distances with fewer update steps and therefore reduced error accumulation.

\subsection{Extended ECHO Results}
\label{app:echo-oor}

\paragraph{Out-of-range splits.}
We construct the out-of-range splits from the original ECHO-Synth train, validation, and test partitions \citep{miglior2025can}, without generating new graphs. The construction differs between node-level and graph-level tasks.

For the node-level tasks, SSSP and eccentricity, every graph of the original training set is used, and all nodes are processed by the model. Supervision is removed from nodes whose target exceeds a task-specific threshold: these nodes still participate in message passing, but do not contribute to the loss computation. The model therefore observes graphs containing long-range structure, but never receives a training signal for targets beyond the threshold. The same masking is applied to the validation set used for model selection. At test time, nodes of the original test set are evaluated separately according to whether their target lies within the training range (IR) or beyond it (OOR).

For the graph-level task, diameter, masking is applied at the level of whole graphs: training and validation graphs whose diameter exceeds the threshold are removed, and test graphs are assigned to IR or OOR according to their diameter.

The thresholds are $21$ hops for SSSP and $30$ for eccentricity and diameter, giving the intervals reported in Table~\ref{tab:echo-synth-oor}. The lower end of each interval is fixed by the construction of ECHO-Synth, whose graphs have diameters between $17$ and $40$.

\paragraph{Training protocol.}
All models use the configuration selected for the corresponding task in Ref. \citep{miglior2025can}, including GHR, and are trained with three seeds. No model performs additional iterations at inference.

\subsection{Extended LRIM Results} 
\label{LRIM-extended}
\begin{table}[ht]
\caption{\textbf{Transfer performance from models trained on LRIM-16-hard to OOD lattices dimensions.} Values represent LogMSE. Baselines values are reported from \cite{mathys2026lrim}. Our GHR used in this OOD task is the same weights trained in the original task reported in \ref{tab:lrim-16-hard}. For each OOD task column, the \textcolor[HTML]{009e74}{\textbf{best}}, \textcolor[HTML]{d55e00}{\textbf{second-best}}, and \textbf{third-best} results are highlighted.}
\label{tab:lrim-transfer-complete}
\begin{center}
\small
\begin{tabular}{lcccc}
\toprule
\textbf{Model} $\downarrow$ & \textbf{32-hard} $\downarrow$ & \textbf{64-hard} $\downarrow$ & \textbf{128-hard} $\downarrow$ & \textbf{256-hard} $\downarrow$ \\
\midrule
GHR-GatedGCN (Ours) & \textcolor[HTML]{009e74}{$\mathbf{-1.086}$ {\scriptsize $\mathbf{\pm 0.014}$}} & \textcolor[HTML]{d55e00}{$\mathbf{-0.825}$ {\scriptsize $\mathbf{\pm 0.023}$}} & \textcolor[HTML]{009e74}{$\mathbf{-0.771}$ {\scriptsize $\mathbf{\pm 0.001}$}} & \textcolor[HTML]{009e74}{$\mathbf{-1.073}$ {\scriptsize $\mathbf{\pm 0.003}$}} \\
GHR-GIN (Ours) & \textcolor[HTML]{d55e00}{$\mathbf{-1.085}$ {\scriptsize $\mathbf{\pm 0.021}$}} & \textcolor[HTML]{009e74}{$\mathbf{-0.843}$ {\scriptsize $\mathbf{\pm 0.030}$}} & \textcolor[HTML]{d55e00}{$\mathbf{-0.770}$ {\scriptsize $\mathbf{\pm 0.011}$}} & \textcolor[HTML]{d55e00}{$\mathbf{-1.066}$ {\scriptsize $\mathbf{\pm 0.033}$}} \\
\cmidrule(r){2-5} 
GIN  & $-1.043$ {\scriptsize $\pm 0.051$} & $-0.774$ {\scriptsize $\pm 0.042$} & $-0.703$ {\scriptsize $\pm 0.041$} & $-0.903$ {\scriptsize $\pm 0.047$}  \\
GatedGCN  & $-1.050$ {\scriptsize $\pm 0.004$} & $-0.781$ {\scriptsize $\pm 0.002$} & $-0.708$ {\scriptsize $\pm 0.003$} & $-0.952$ {\scriptsize $\pm 0.005$} \\
GatedGCN-VN$_G$  & $-1.054$ {\scriptsize $\pm 0.006$} & $-0.788$ {\scriptsize $\pm 0.004$} & $-0.716$ {\scriptsize $\pm 0.005$} & $\mathbf{-0.968}$ {\scriptsize $\mathbf{\pm 0.008}$} \\
GPS-Base  & $\mathbf{-1.057}$ {\scriptsize $\mathbf{\pm 0.000}$} & $\mathbf{-0.790}$ {\scriptsize $\mathbf{\pm 0.001}$} & $\mathbf{-0.719}$ {\scriptsize $\mathbf{\pm 0.001}$} & OOM \\
GPS-RWSE  & $\mathbf{-1.057}$ {\scriptsize $\mathbf{\pm 0.005}$} & $\mathbf{-0.790}$ {\scriptsize $\mathbf{\pm 0.004}$} & $-0.716$ {\scriptsize $\pm 0.002$} & OOM \\
GPS-LapPE  & $-1.053$ {\scriptsize $\pm 0.006$} & $-0.785$ {\scriptsize $\pm 0.005$} & $-0.716$ {\scriptsize $\pm 0.005$} & OOM \\
\bottomrule
\end{tabular}
\end{center}
\end{table}

\subsection{LRGB Results} 
\label{LRGB-appendix}

We evaluate GHR on two tasks from the Long Range Graph Benchmark \citep{LRGB}, PascalVOC-SP and Peptides-struct, using the three standard backbones. Baselines are the re-tuned configurations of these operators under Ref. \citep{tonshoff2024where}. Each GHR variant uses the same message-passing operator as its flat counterpart, so the comparison isolates the framework rather than the backbone.
On PascalVOC-SP, GHR improves over the corresponding flat model for all three backbones. On Peptides-struct, all models are close in performances with GHR matching the flat baselines. This is consistent with prior analyses showing that Peptides-struct depends weakly on long-range interaction and that simply adjusting the readout layer is what makes every model achieve similar performances \citep{tonshoff2024where, bamberger2025on}, and with the observation that performance on this task is agnostic to the specific architecture once hyperparameters are tuned. We therefore include LRGB as a demonstration that the framework transfers to widely used benchmarks.

\begin{table}[ht]
\caption{\textbf{Results on the Pascal VOC and Peptides-struct Tasks.} Baseline performance metrics are reported from \citep{tonshoff2024where}.}
\label{tab:pascal-peptides}
\centering
\small
\setlength{\tabcolsep}{8pt}
\begin{tabular}{lcc}
\toprule
\textbf{Model} & \textbf{Pascal VOC} (Test F1 $\uparrow$) & \textbf{Peptides-struct} (Test MAE $\downarrow$) \\
\midrule
GCN & $0.2078$ {\scriptsize $\pm 0.0031$} & $0.2460$ {\scriptsize $\pm 0.0007$} \\
GHR-GCN (Ours) & $0.2198$ {\scriptsize $\pm 0.0030$} & $0.2454$ {\scriptsize $\pm 0.0013$} \\
\midrule
GINE & $0.2718$ {\scriptsize $\pm 0.0054$} & $0.2473$ {\scriptsize $\pm 0.0017$} \\
GHR-GINE (Ours) & $0.3001$ {\scriptsize $\pm 0.0054$} & $0.2460$ {\scriptsize $\pm 0.0015$} \\
\midrule
GatedGCN & $0.3880$ {\scriptsize $\pm 0.0040$} & $0.2477$ {\scriptsize $\pm 0.0009$} \\
GHR-GatedGCN (Ours) & $0.4017$ {\scriptsize $\pm 0.0051$} & $0.2469$ {\scriptsize $\pm 0.0011$} \\
\bottomrule
\end{tabular}
\end{table}

\subsection{Training Configurations}
\label{app:hyperparameters}

Table~\ref{tab:echo_hyperparameters} to Table~\ref{tab:LRGB-hyperparameters} report the hyperparameter best values selected for each task across the different GHR implementation we considered. Best values were identified through an hyperparameter tuning on the validation set.
We remark that best configuration are shared across backbones to isolate the Message-Passing contribution.

\begin{table}[h]
\caption{\textbf{Hyperparameter Configurations of GHR-GINE and GHR-GCN for ECHO Benchmarks.}}
\label{tab:echo_hyperparameters}
\begin{center}
\small
\begin{tabular}{lccccc}
\toprule
\textbf{Hyperparameter} & \textbf{Charge} & \textbf{Energy} & \textbf{Diam} & \textbf{ECC} & \textbf{SSSP} \\
\midrule
Hidden Dimension ($d$) & $32$ & $32$ & $32$ & $32$ & $32$ \\
Low-Level Steps ($L$) & $3$ & $3$ & $2$ & $2$ & $6$ \\
High-Level Steps ($H$) & $3$ & $3$ & $2$ & $3$ & $3$ \\
Learning Rate & $3.4 \times 10^{-4}$ & $3.0 \times 10^{-4}$ & $1.8 \times 10^{-3}$ & $1.0 \times 10^{-3}$ & $1.0 \times 10^{-3}$ \\
Batch Size & $128$ & $128$ & $128$ & $256$ & $128$ \\
\# of Global Recurrent Steps  & $1$ & $1$ & $4$ & $4$ & $2$ \\
Activation of MLP (GINE) & SiLU & SiLU & SiLU & SwiGLU & SwiGLU  \\

\bottomrule
\end{tabular}
\end{center}
\end{table}

\begin{table}[h]
\caption{\textbf{Hyperparameter Configurations of GHR-GatedGINE and GHR-GCN for LRIM.}}
\label{tab:LRIM_hyperparameters}
\begin{center}
\small
\begin{tabular}{lcc}
\toprule
\textbf{Hyperparameter} & \textbf{16-Hard} & \textbf{32-Hard} \\
\midrule
Hidden Dimension ($d$) & $300$ & $256$ \\
Low-Level Steps ($L$) & $3$ & $3$ \\
High-Level Steps ($H$) & $3$ & $6$ \\
\# of Global Recurrent Steps & $1$ & $1$ \\
Learning Rate & $3 \times 10^{-4}$ & $3 \times 10^{-4}$ \\
Activation of MLP (GINE) & SiLU & SiLU \\
\bottomrule
\end{tabular}
\end{center}
\end{table}

\begin{table}[h]
\caption{\textbf{Hyperparameter Configurations of GHR-GINE, GHR-GCN, GHR-GatedGCN for LRGB.} We remark that in LRGB we had to vary \texttt{hidden\_dimension} in order to adhere to the 500k parameter budget with the 3 different backbones, everything else is shared between them.}
\label{tab:LRGB-hyperparameters}
\begin{center}
\small
\begin{tabular}{lcc}
\toprule
\textbf{Hyperparameter} & \textbf{Pascal VOC} & \textbf{Peptides-struct} \\
\midrule
Hidden Dimension GINE ($d$) & $300$ & $300$ \\
Hidden Dimension GCN & $350$  & $350$\\
Hidden Dimension GatedGCN & $220$  & $300$  \\
Low-Level Steps ($L$) & $3$ & $3$ \\
High-Level Steps ($H$) & $4$ & $3$ \\
\# of Global Recurrent Steps & $1$ & $1$ \\
Learning Rate & $2.4 \times 10^{-3}$ & $ 1\times 10^{-3}$ \\
Dropout & $0.2$ & $0.1$  \\
Weight decay & $1\times 10^{-4}$ & 0.0  \\
Activation of MLP (GINE) & SiLU & SiLU \\

\bottomrule
\end{tabular}
\end{center}
\end{table}

Tables \ref{tab:echo_hyperparameters}--\ref{tab:LRGB-hyperparameters} list the hyperparameter configurations for all benchmarks.
We do not use dropout or weight decay. At the parameter budgets used, including the larger budget for LRIM, we observed no overfitting on validation across benchmarks, making regularization unnecessary.

\section{Pooling Operators: Details and Ablations}
\label{pooling-examples}

\begin{table}[h]
\caption{\textbf{Pooling Operator and Cluster Aggregation per Task.} Graclus performs greedy edge-contraction matching; geometric block pooling deterministically partitions a regular lattice into $2 \times 2$ sub-grids.}
\label{tab:pooling-per-task}
\begin{center}
\small
\begin{tabular}{lcc}
\toprule
\textbf{Task} & \textbf{Pool Operator} & \textbf{Cluster Aggregation} \\
\midrule
RGG SSSP & Graclus (2 iter) & max \\
ECHO-Synth (SSSP, Ecc, Diam) & Graclus (2 iter) & max \\
ECHO-Chem (Charge, Energy) & Graclus (1 iter) & sum \\
LRGB (Peptides-struct, PascalVOC) & Graclus (2 iter) & mean \\
LRIM (all variants) & Geometric $2 \times 2$ block & sum \\

\bottomrule
\end{tabular}
\end{center}
\end{table}


\paragraph{Graclus.}
In general we adopt Graclus~\citep{dhillon2007weighted}, which computes a greedy edge matching in a single pass. Each unmatched node $u \in V$ is paired with an unmatched neighbor $w \in \mathcal{N}_u$, preferring neighbors of low degree, and the pair is contracted into a super-node $v' \in V'$, approximately halving $|V|$. Edges between members of different super-nodes are contracted to form $E'$. Within $\mathrm{Pool}$, the features of the nodes in a cluster are combined by the cluster aggregation operator, $\max$ for the algorithmic tasks and $\mathrm{sum}$ for the chemical and physical ones (Table~\ref{tab:pooling-per-task}). The inverse operator $\mathrm{Unpool}(G')$ broadcasts the feature of each super-node back to its constituent nodes. Graclus can be applied repeatedly, approximately halving the number of nodes at each pass; the number of passes $k$ is a hyperparameter that sets the pooling ratio $|V_H|/|V_L|$.

\paragraph{Geometric block renormalization.}
For periodic square lattices, we apply a deterministic block partition. Given a block size $b$, each node with lattice coordinates $(x_i, y_i)$ is assigned to the super-node of the non-overlapping $b \times b$ sub-grid containing it. The coarse edge set $E'$ is obtained by mapping lattice edges to their clusters and removing self-loops, and node features are aggregated by $\mathrm{sum}$. $\mathrm{Unpool}(G')$ broadcasts each super-node's feature back to its $b^2$ nodes. The lattice coordinates are used only to determine the cluster assignment, no positional information is provided to the model as a node feature, and, since coarse edges connect adjacent blocks only, they carry no geometric attribute.

\paragraph{Random coarsening.}
As a control, we also consider a random partition of $V$ into clusters of equal size, chosen to match the compression ratio of the default coarsening. Nodes are assigned to clusters uniformly at random, without regard to adjacency, so a cluster may contain nodes that are far apart in $G_L$. The coarse graph is then constructed accordingly to the other operators: two super-nodes are connected whenever any of their members are adjacent, so random coarsening is still a graph quotient map and satisfies the condition of Section~\ref{background}. The coarsening is computed once per graph before training and kept fixed.

\paragraph{Coarse edges.}
In all cases, fine edges whose endpoints fall in the same cluster become self-loops and are removed, so information carried only by intra-cluster edges reaches $G_H$ through node-feature pooling alone. Edges between different clusters are merged into a single coarse edge, whose features depend on the dataset. On ECHO, the features of all merged fine edges are summed, and the size of the source cluster is appended. On RGG, whose fine edges carry no information, coarse edges are assigned a constant connectivity feature together with the source cluster size. On LRIM, coarse edges produced by graclus or random matching carry the number of merged fine edges, which instead is constant when using block renormalization.

\paragraph{Relation to the flat recurrent model.}
The hierarchy does not restrict the class of functions GHR can represent. As detailed in Section~\ref{section:reasoning_step}, the high-level state enters the low-level update only as an addition and through the learnable map $W''_L$. For $W''_L = 0$, GHR reduces to a flat recurrent MPNN with the same number of low-level message-passing steps, so its function class contains that of the flat model. An inadequate coarsening therefore adds computation without reducing expressivity relative to the flat model.
When the coarse abstraction aids propagation, the high-level stream improves performance, as shown in Section~\ref{experiments} and Appendix~\ref{app:pooling-ablation}.

\subsection{Pooling and Coarsening Ablation}
\label{app:pooling-ablation}

The previous section shows that GHR contains the flat recurrent model as a special case, so an unsuitable coarsening should degrade performance towards that of the flat model.
We test this, and perform ablations on coarsening choices, on ECHO-SSSP and LRIM-16-hard.

\paragraph{LRIM-16-hard.}
We evaluate GHR-GatedGCN with $R = 1$, $T_H = 3$, $T_L = 3$, and hidden dimension $256$, matching Table~\ref{tab:lrim-16-hard}. The $9$ low-level steps fall below the lattice diameter of $16$. At an approximate $1/4$ compression ratio, we compare $2 \times 2$ blocks, Graclus ($k = 2$), a random four-node partition, and a flat baseline without the coarse stream, alongside $4 \times 4$ blocks ($b = 4$).
As shown in Table~\ref{tab:pooling-ablation-lrim}, metric values range from $-2.29$ (flat GHR baseline) to $-4.70$ ($2 \times 2$ blocks), establishing that the coarse stream drives the performance gain. Graclus and random partitioning match each other and capture most of this improvement despite ignoring lattice regularity. Random clustering mixes distant nodes into the same cluster, which is precisely what GHR otherwise avoids. On LRIM, the Hamiltonian couples all spin pairs, making targets dependent on global lattice aggregates so Random clustering is producing a denser coarse graph (average degree $14.2$ vs.\ $4$ for blocks) that could capture these global interactions.  On ECHO-SSSP, where targets depend on exact distances, random clustering instead helps little beyond the reach of the fine stream (Table~\ref{tab:pooling-ablation-echo}). Although $4 \times 4$ blocks degrade performance relative to $2 \times 2$, the resulting model still outperforms most baselines of Table~\ref{tab:lrim-16-hard}.

Block pooling incorporates structural knowledge without providing positional features or geometric edge attributes. For periodic grids, $b \times b$ clusters derive strictly from topology and remain fixed prior to training, unlike the explicit Laplacian positional encodings fed to GPS-LapPE. Coarsening improves performance across all partition choices, this demonstrates that the hierarchy itself is the primary mechanism for information propagation, whereas adjusting the coarsening to graph geometry gives the remaining margin to establish state-of-the-art results.

\begin{table}[ht]
\caption{\textbf{Coarsening ablation on LRIM-16-hard.} GHR-GatedGCN with $R = 1$, $T_H = 3$, $T_L = 3$ and hidden dimension $256$, varying only the coarsening. The compression ratio is $|V_H|/|V_L|$. Values represent LogMSE ($\downarrow$), averaged over three seeds.}
\label{tab:pooling-ablation-lrim}
\begin{center}
\footnotesize
\setlength{\tabcolsep}{8pt}
\setlength{\aboverulesep}{0pt}
\setlength{\belowrulesep}{0pt}
\renewcommand{\arraystretch}{1.05}
\begin{tabular}{l c c c}
\toprule
\textbf{Coarsening} & \textbf{Compression ratio} & \textbf{LogMSE} $\downarrow$ & Coarse Graph Degree \\
\midrule
Block $2 \times 2$ & $1/4$ & $-4.701_{\pm 0.031}$ & $4.0_{\pm 0.0}$ \\
Block $4 \times 4$ & $1/16$ & $-4.098_{\pm 0.012}$ & $4.0_{\pm 0.0}$ \\
Graclus ($k = 2$) & $\approx 1/4$ & $-3.630_{\pm 0.015}$ & $5.4_{\pm 0.1}$ \\
Random partition & $1/4$ & $-3.669_{\pm 0.009}$ & $14.2_{\pm 0.2}$ \\
\midrule
No coarse stream & -- & $-2.294_{\pm 0.020}$ & - \\
\bottomrule
\end{tabular}
\end{center}
\end{table}

\paragraph{ECHO-SSSP.}
We use GHR-GINE with $R = 2$, $T_H = 3$, $T_L = 3$ and hidden dimension $32$. This depth is reduced relative to Table~\ref{tab:combined-results}, so that the fine stream alone, with $R\,T_H\,T_L = 18$ low-level steps, cannot span the graphs and the contribution of the coarse stream can be isolated. We vary the compression ratio with graclus using $k \in \{1, 2, 3, 5, 10\}$ passes, compare graclus with a random partition at the ratio of the default configuration ($k = 2$), and report GHR without the coarse stream.
Since most test nodes lie close to the source, aggregated MAE is dominated by distances the low-level stream already covers. 

We therefore report the error separately for nodes within the receptive field of the low-level stream and beyond it in order to explicitly show the hierarchy contribution (Table~\ref{tab:pooling-ablation-echo}). 
Within $18$ hops of the source, where $79\%$ of the test nodes lie, the flat recurrence defined by GHR is already accurate, as we saw in Section~\ref{sec:Ablations}. Beyond this range, the error of the flat model rises sharply, whereas graclus with moderate compression ($k \le 2$) reduces it by nearly a factor of three. Stronger compression removes this benefit, and at $k = 10$, where the coarse graph collapses to a single node, the model's error is compatible with that of the flat recurrent model within one standard deviation.
A random partition, which performed comparably to graclus on LRIM, helps less in this case: beyond the reach of the low-level stream it reduces the error of the flat model in the $19-40$ hops regime only from $3.04$ to $2.49$, being less effective than graclus. Shortest-path targets depend on exact distances, and clusters that group distant nodes randomly provide shortcuts that do not necessarily aid the model and this results in a reduced performance with respect to a partition like graclus. 

\begin{table}[ht]
\caption{\textbf{Coarsening ablation on ECHO-SSSP, by distance regime.} GHR-GINE with $R = 2$, $T_H = 3$, $T_L = 3$, hidden dimension $32$, varying only the coarsening. Values are test MAE ($\downarrow$).}
\label{tab:pooling-ablation-echo}
\begin{center}
\footnotesize
\setlength{\tabcolsep}{8pt}
\setlength{\aboverulesep}{0pt}
\setlength{\belowrulesep}{0pt}
\renewcommand{\arraystretch}{1.05}
\begin{tabular}{l c c c}
\toprule
\textbf{Coarsening} & \textbf{Hops 0-18} & \textbf{Hops 19-40} & \textbf{Overall} \\
\midrule
Graclus ($k = 1$)  & $0.348_{\pm 0.023}$ & $1.130_{\pm 0.060}$ & $0.515_{\pm 0.022}$ \\
Graclus ($k = 2$)  & $0.218_{\pm 0.036}$ & $1.113_{\pm 0.106}$ & $0.410_{\pm 0.036}$ \\
Graclus ($k = 3$)  & $0.444_{\pm 0.073}$ & $1.370_{\pm 0.066}$ & $0.642_{\pm 0.059}$ \\
Graclus ($k = 5$)  & $0.455_{\pm 0.082}$ & $1.479_{\pm 0.115}$ & $0.674_{\pm 0.069}$ \\
Graclus ($k = 10$) & $0.305_{\pm 0.010}$ & $3.060_{\pm 0.246}$ & $0.895_{\pm 0.053}$ \\
Random partition   & $0.364_{\pm 0.008}$ & $2.490_{\pm 0.139}$ & $0.819_{\pm 0.030}$ \\
\midrule
No coarse stream   & $0.305_{\pm 0.022}$ & $3.040_{\pm 0.006}$ & $0.890_{\pm 0.017}$ \\
\bottomrule
\end{tabular}
\end{center}
\end{table}

Overall, these results also better characterize the regimes in which the coarsening procedure is limited or fails in aiding the model. When the coarse graph does not preserve task-relevant topological structure, is obtained via excessive compression or clusters which are too coarse for the required resolution, performance degrades towards that of the model without the coarse stream.
This behavior is consistent with the structure of the model: since the coarse stream enters the low-level update only through $W''_L$, setting $W''_L = 0$ recovers the flat recurrent model (Appendix~\ref{pooling-examples}), so a not useful coarsening can at worst be disregarded rather than constrain the function the model represents.

\section{Limitations}
\label{sec:limitations}

Recent work has made substantial progress toward more principled benchmarks for long-range graph learning \citep{dwivedi2021generalization, bamberger2025on, miglior2025can, mathys2026lrim}. Nevertheless, the availability of large-scale, realistic benchmarks in which interaction range can be systematically varied independently of other distributional factors remains limited.
Our experiments combine controlled settings that isolate interaction range with established benchmarks spanning synthetic and real-world domains. Further developing large-scale realistic datasets and tasks with explicitly controllable interaction ranges represents an important direction for future work.

\clearpage

\end{document}